\documentclass[conference]{IEEEtran}
\IEEEoverridecommandlockouts
\usepackage{placeins}
\usepackage{amsmath,amssymb,amsthm,mathtools}
\usepackage{booktabs}
\usepackage{graphicx}
\usepackage{float}
\usepackage[dvipsnames]{xcolor}
\usepackage{tikz}
\usetikzlibrary{arrows.meta,positioning}
\usepackage[noend]{algpseudocode}
\usepackage{tcolorbox}
\tcbuselibrary{skins}
\usepackage{cite}
\usepackage{enumitem}
\usepackage[hypertexnames=false,colorlinks=true,linkcolor=blue!55!black,citecolor=blue!55!black,urlcolor=blue!55!black]{hyperref}
\newcommand{\R}{\mathbb{R}}
\newcommand{\E}{\mathbb{E}}
\newcommand{\diag}{\operatorname{diag}}
\newcommand{\blkdiag}{\operatorname{blkdiag}}
\newcommand{\half}{\tfrac{1}{2}}
\newcommand{\ones}{\mathbf{1}}
\DeclareMathOperator*{\argmin}{arg\,min}

\newtheorem{remark}{Remark}

\newcommand{\AverageSmallMatrix}[1]{\bigl[\begin{smallmatrix}#1\end{smallmatrix}\bigr]}
\newcommand{\newalgcomment}[2]{\hfill{\footnotesize #1~\ref{#2}}}
\makeatletter
\newcommand{\customlabel}[2]{\protected@write\@auxout{}{\string\newlabel{#1}{{#2}{\thepage}{#2}{#1}{}}}\hypertarget{#1}{#2}}
\makeatother
\tcbset{
diffmpcOuterFull/.style={
    enhanced,
    before skip=0pt,
    after skip=0pt,
    coltitle=black,
    boxrule=3pt,
    colframe=gray!16,
    colback=gray!8,
    colbacktitle=gray!18,
    boxsep=0pt,
    left=6pt,
    right=6pt,
    top=2pt,
    bottom=2pt,
},
diffmpcForwardFull/.style={
    enhanced,
    coltitle=black,
    boxrule=3pt,
    colframe=ForestGreen!10,
    colback=ForestGreen!4,
    colbacktitle=ForestGreen!10,
    boxsep=0pt,
    left=6pt,
    right=6pt,
    top=2pt,
    bottom=2pt,
},
diffmpcADMMFull/.style={
    enhanced,
    before skip=0pt,
    after skip=4pt,
    coltitle=black,
    boxrule=3pt,
    colframe=ForestGreen!10,
    colback=ForestGreen!4,
    colbacktitle=ForestGreen!10,
    boxsep=0pt,
    left=6pt,
    right=6pt,
    top=2pt,
    bottom=2pt,
}
}
\definecolor{DiffMPCPurple}{HTML}{F2CCFF}
\definecolor{DiffMPCPurpleBorder}{HTML}{8200D8}
\tcbset{
diffmpcBackwardFull/.style={
    enhanced,
    before skip=0pt,
    after skip=0pt,
    coltitle=black,
    boxrule=3pt,
    colframe=DiffMPCPurpleBorder!12,
    colback=DiffMPCPurpleBorder!4,
    colbacktitle=DiffMPCPurpleBorder!10,
    boxsep=0pt,
    left=6pt,
    right=6pt,
    top=2pt,
    bottom=2pt,
}
}
\definecolor{DiffMPCBlueBorder}{HTML}{0057D8}
\tcbset{
diffmpcTrainFull/.style={
    enhanced,
    before skip=0pt,
    after skip=0pt,
    coltitle=black,
    boxrule=3pt,
    colframe=DiffMPCBlueBorder!12,
    colback=DiffMPCBlueBorder!4,
    colbacktitle=DiffMPCBlueBorder!10,
    boxsep=0pt,
    left=6pt,
    right=6pt,
    top=2pt,
    bottom=2pt,
}
}

\makeatletter
\newcommand{\compactsubsection}{\@startsection{subsection}{2}{\z@}
  {5pt}{3pt}{\normalfont\normalsize\itshape}}
\makeatother
\definecolor{dartGreen}{RGB}{0, 105, 62}

\newcommand{\solverName}{SmoothedTurboMPC}

\title{\LARGE \bf Efficient Learning of Differentiable MPC Policies on the GPU\\via Smoothed ADMM and First-Order Reinforcement Learning}

\author{\IEEEauthorblockN{Jianghan Zhang\textsuperscript{1}, Roland Schwan\textsuperscript{1}, Gabriel Bravo-Palacios\textsuperscript{1}, John Zhang\textsuperscript{2}\\
Thomas Lew\textsuperscript{3}, and Brian Plancher\textsuperscript{1}}\thanks{\textsuperscript{1}Jianghan Zhang, Roland Schwan, Gabriel Bravo-Palacios, Brian Plancher are with Dartmouth College, Hanover, NH, USA, \href{mailto:jianghan.zhang.gr@dartmouth.edu,roland.schwan@dartmouth.edu,gbravo@dartmouth.edu,plancher@dartmouth.edu}{\texttt{\{jianghan.zhang.gr, roland.schwan, gbravo, plancher\}@dartmouth.edu}}.}\thanks{\textsuperscript{2}John Zhang is with the Massachusetts Institute of Technology, Cambridge, MA, USA, \href{mailto:jzhang3@mit.edu}{\texttt{jzhang3@mit.edu}}.}\thanks{\textsuperscript{3}Thomas Lew is with the Toyota Research Institute, Los Altos, CA, USA, \href{mailto:thomas.lew@tri.global}{\texttt{thomas.lew@tri.global}}.}\thanks{Toyota Research Institute provided funds to support this work, but this article solely reflects the opinions and conclusions of its authors.}
\thanks{This project was supported by the National Science Foundation (Awards 2411369, 2535096). The opinions, findings, conclusions, and recommendations are those of the authors and do not necessarily reflect those of the funding organizations.}
}

\hypersetup{
  pdftitle={Efficient Learning of Differentiable MPC Policies on the GPU via Smoothed ADMM and First-Order Reinforcement Learning},
  pdfauthor={Jianghan Zhang, Roland Schwan, Gabriel Bravo-Palacios, John Zhang, Thomas Lew, Brian Plancher}
}

\begin{document}
\maketitle
\thispagestyle{empty}
\pagestyle{empty}

\begin{abstract}
    Differentiable model predictive control (MPC) enables structured, constraint-aware policies to be optimized end-to-end.
However, learning MPC policies faces two bottlenecks:  (1) nonsmooth sensitivities at active-set transitions and (2) expensive policy evaluation, since each requires an optimization solve.
We first introduce \solverName{}, a GPU-parallel differentiable MPC solver combining sequential quadratic programming (SQP) with a log-barrier-smoothed alternating direction method of multipliers (ADMM) inner solver. Its design enables smooth gradients, exact sensitivities at convergence, and efficient execution on GPUs. 
Second, to efficiently train MPC policies, we pair our solver with first-order reinforcement learning (FORL) using Short-Horizon Actor-Critic (SHAC), which reuses rollout MPC solutions instead of repeatedly re-evaluating them. Comparisons with state of the art solvers show that the approach achieves up to $4.4\times$ higher end-to-end training throughput.
On a vehicle racing task, our method achieves higher returns than MPC PPO with a $46\times$ speedup, training on 3.3 million transitions in only 7 minutes.
Finally, we use the framework to train a more reliable MPC racing policy via RL with domain randomization, and validate it on a full-scale race car. We integrate our advances into the open-source TurboMPC solver.

\end{abstract}

\begin{flushleft}
\footnotesize
\textbf{Code:} \url{https://github.com/ToyotaResearchInstitute/turbompc}\\
\textbf{Video:} \url{https://www.youtube.com/watch?v=aUvfUL85Mcg}
\end{flushleft}

\section{Introduction}
\label{sec:intro}

Model predictive control (MPC) is an attractive structured policy class for learning-based control. 
By solving optimization problems in real time,  
MPC incorporates model knowledge, plans over a finite horizon, and explicitly handles state and input constraints.
Learning can adapt MPC cost and model parameters to task distributions while retaining this structure, with potential benefits for adaptation, generalization, and robustness~\cite{gros2020datadriven,song2022policysearch,zarrouki2021rlwmpc,frohlich2022contextual}. 
Differentiable MPC further enables these policies to be optimized end-to-end by combining an MPC solve (the forward pass) with sensitivities of its solution map to policy parameters (the backward pass)~\cite{optnet,amos2018diffmpc,difftunempc,frey2025diffnmpc,adabag2026diffmpc}.

\begin{figure}[!t]
  \centering
  \includegraphics[width=0.97\columnwidth]{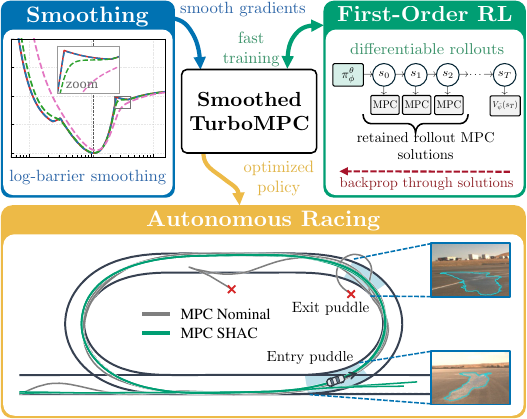}
  \vspace{-5pt}
  \caption{\textbf{Overview.} 
  (Top) The proposed differentiable MPC solver uses log-barrier regularization to return smooth gradients. A first-order RL framework is co-designed to enable efficient RL training of MPC policies end-to-end. (Bottom) The resulting framework with domain randomization enables reliable vehicle racing of an MPC controller with a neural network cost map.}
  \label{fig:intro_overview}
  \vspace{-8pt}
\end{figure}

Two obstacles limit efficient learning with differentiable MPC.
\textbf{1) Nonsmooth sensitivities.}  
The solution of MPC's underlying constrained nonlinear program (NLP) becomes nonsmooth as the constraint active set changes, resulting in discontinuous or noisy gradients near constraint boundaries~\cite{tracy2024ipm,frey2025diffnmpc,jahncke2026diffwmpc}, hindering learning.
Existing methods primarily address this via log-barrier interior-point methods (IPMs)~\cite{tracy2024ipm,frey2025diffnmpc,difftunempc}. \textbf{2) Expensive policy evaluation.}
Evaluating a neural network  policy is cheap, whereas evaluating an MPC policy requires an optimization solve. RL-MPC methods that repeatedly re-evaluate the policy on collected data reduce environment interactions, but overall increase training time because MPC solves can be expensive.

In this work, we co-design a differentiable MPC solver with its learning pipeline (Figure~\ref{fig:intro_overview}) to address these bottlenecks: 

\textbf{1) 
A log-barrier-smoothed SQP-ADMM differentiable solver.}
We introduce \solverName{}, a GPU-parallel differentiable MPC solver that builds on TurboMPC~\cite{turbompc} and returns smooth gradients, 
\solverName{} replaces its projection-based inequality update with a log-barrier proximal update, enabling efficient solves while enforcing perturbed complementarity.
To the best of our knowledge, this is the first integration of log-barrier smoothing with ADMM~\cite{boyd2011admm} for differentiable nonlinear MPC, providing a smooth implicit sensitivity model throughout SQP and exact smoothed-NLP sensitivities at convergence through implicit differentiation.

\textbf{2) A computation-efficient RL pipeline for differentiable MPC.}
We pair \solverName{} with solve-efficient FORL using Short-Horizon Actor-Critic (SHAC)~\cite{xu2022shac,amigo2025decoupled}.
Our pipeline retains rollout MPC solutions for a single first-order policy update rather than repeatedly re-evaluating the policy on collected states.

Together, these choices enable efficient MPC policy learning.
Across differentiable MPC benchmarks, \solverName{} achieves up to $4.4\times$ higher end-to-end training throughput.
In simulated autonomous racing, MPC-SHAC achieves higher returns than PPO and a $46\times$ wall-clock speedup, training 3.3 million steps in only 7.0 minutes.
We deploy the learned policy on a full-scale Lexus LC500 in closed-course racing with a puddle at a corner's entry or exit. Both controllers approach the puddle at comparable speeds, the learned policy completes all four trials, including 12 puddle passes, while the nominal MPC spins out in all four trials.

\section{Background}
\label{sec:background}

\subsection{Differentiable Optimization through Implicit Differentiation}
\label{sec:bg-ift}
MPC repeatedly solves optimization problems to compute control inputs actuating the system.  Thus, learning with MPC policies requires a method for computing the gradients of the solutions to optimization problems, which we describe next. 

At each control step, MPC solves a nonlinear program (NLP) of the form:
\begin{equation}
\label{eq:nlp}
\min_{x}\ \ell(x;\theta)
\quad\text{s.t.}\quad f(x;\theta) = 0,\ \ g(x;\theta) \le h(\theta),
\end{equation}
where $x\in\R^n$ are the decision variables, $f(x;\theta)\in\R^m$ and $g(x;\theta),h(\theta)\in\R^p$ define the constraints, and $\theta\in\R^{n_\theta}$ denotes problem parameters (e.g., cost weights).
For multipliers $(y_f, y_g)$, the Lagrangian of the NLP is:
\begin{equation}
\label{eq:lagrangian}
\mathcal L := \ell + y_f^\top f + y_g^\top (g - h).
\end{equation}
Introducing slack variables $\nu:=h-g(x)$, and defining $w:=(x,y_f,y_g)$ and the Jacobians $C:=\partial f/\partial x$ and $G:=\partial g/\partial x$, the Karush-Kuhn-Tucker (KKT) conditions~\cite{nocedalwright}  of the NLP are:
\begin{equation}
\label{eq:nlp-kkt}
F^{\mathrm{NLP}}(w;\theta):=
\begin{bmatrix}
\nabla_x\ell+C^\top y_f+G^\top y_g\\[1pt]
f\\[1pt]
\nu\circ y_g
\end{bmatrix}=0,
\end{equation}
with $\nu \ge0$, $y_g\ge0$, and the elementwise product $\circ$. 

The solution $w^\star(\theta)$ of the NLP depends on the parameters $\theta$. Implicit differentiation techniques \cite{optnet,difftunempc,amos2018diffmpc,frey2025diffnmpc,tracy2024ipm,adabag2026diffmpc}  compute the gradient of this solution by differentiating the KKT conditions \eqref{eq:nlp-kkt}.  
Writing \eqref{eq:nlp-kkt} as a root-finding problem $F^{\mathrm{NLP}}(w;\theta) = 0$,
let $\mathcal{M} := \partial F^{\mathrm{NLP}}/\partial w$, if $\mathcal{M}$ is nonsingular at $(w^\star, \theta)$, the implicit function theorem (IFT) yields a locally unique $C^1$ map $\theta \mapsto w(\theta)$ with
\begin{equation}
\label{eq:ift}
\frac{\partial w}{\partial\theta} = -\mathcal{M}^{-1}\frac{\partial F^{\mathrm{NLP}}}{\partial\theta}.
\end{equation}

Learning introduces an outer scalar task loss $L:\R^n\rightarrow\R$, evaluated at the NLP solution, yielding the bilevel problem:
\begin{equation}
\label{eq:bilevel}
\min_{\theta}\ L\bigl(x^\star(\theta)\bigr)
\quad\text{s.t.}\quad
x^\star(\theta)\ \text{solves \eqref{eq:nlp}}.
\end{equation}
Assuming $L$ depends on $\theta$ only through $x^\star(\theta)$, its parameter gradient can be computed without explicitly forming $\partial w/\partial\theta$.
Let $\mathcal H:=\nabla^2_{xx}\mathcal L$ denote the Lagrangian Hessian
evaluated at $w^\star(\theta)$. We define the adjoint variables
$\lambda:=(\lambda_x,\lambda_f,\lambda_g)$ by:
\begin{equation}
\label{eq:adjoint}
\underbrace{\begin{bmatrix}
\mathcal H & C^\top & -G^\top Y\\
C & 0 & 0\\
G & 0 & D
\end{bmatrix}}_{\displaystyle \mathcal{M}^\top}
\begin{bmatrix}
\lambda_x\\ \lambda_f\\ \lambda_g
\end{bmatrix}
=
-
\begin{bmatrix}
\nabla_x L\\0\\0
\end{bmatrix},
\end{equation}
with $D:=\diag(\nu)$, $Y:=\diag(y_g)$. The outer gradient is thus:
\begin{equation}
\label{eq:outer-gradient}
\nabla_\theta L
=
\Bigl(\frac{\partial F^{\mathrm{NLP}}}{\partial\theta}\Bigr)^{\!\top}
\lambda .
\end{equation}

\subsection{Active-Set Changes and Barrier Smoothing}
\label{sec:bg-nonsmooth}

At an active-set transition where strict complementarity fails for constraint $i$, both $\nu_i=0$ and $y_{g,i}=0$. The column of $\mathcal{M}^\top$ in~\eqref{eq:adjoint} associated with the $i$th inequality multiplier then vanishes, making the KKT Jacobian singular and the standard IFT sensitivity undefined~\cite{optnet,frey2025diffnmpc}. This can produce discontinuous solution sensitivities as constraints enter or leave the active set.
Log-barrier smoothing avoids this by replacing hard complementarity, $\nu \circ y_g=0$,
with a perturbed condition~\cite{tracy2024ipm,frey2025diffnmpc}:
\begin{equation}
\label{eq:perturbed-complementarity}
    \nu\circ y_g=\kappa\mathbf{1},
    \qquad \nu>0,\quad y_g>0,\quad \kappa>0 .
\end{equation}
The barrier parameter $\kappa$ trades approximation of the original NLP for smoothness. As $\kappa\rightarrow0^+$, the perturbed problem approaches the hard-constrained problem, while $\kappa>0$ prevents $\nu_i$ and $y_{g,i}$ from simultaneously vanishing. 
\begin{figure}[!t]
  \centering
  \includegraphics[width=0.485\columnwidth]
    {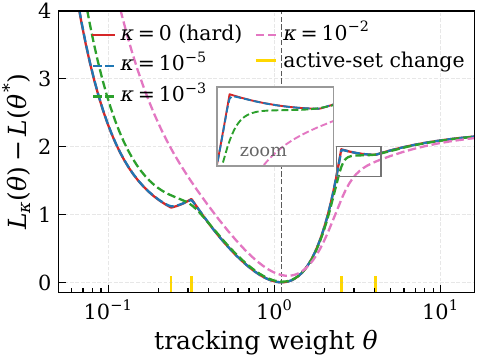}\hfill
  \includegraphics[width=0.485\columnwidth]
    {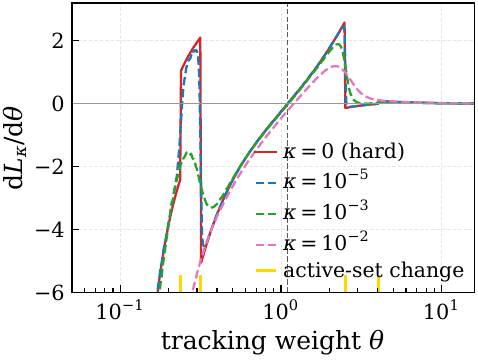}\\[-2pt]
  \makebox[0.485\columnwidth]{\scriptsize (a) Smoothed outer loss}\hfill
  \makebox[0.485\columnwidth]{\scriptsize (b) Smoothed gradient}\\[2pt]
  \includegraphics[width=0.485\columnwidth]
    {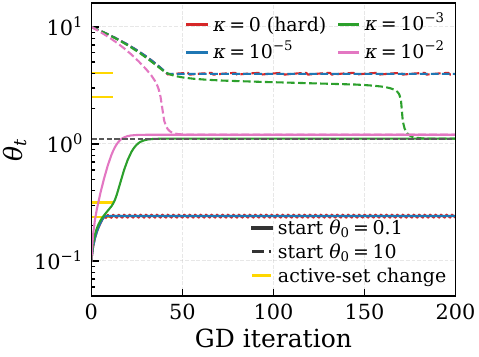}\hfill
  \includegraphics[width=0.485\columnwidth]
    {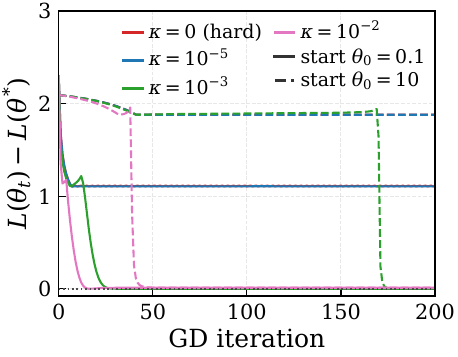}\\[-2pt]
  \makebox[0.485\columnwidth]{\scriptsize (c) GD iterates}\hfill
  \makebox[0.485\columnwidth]{\scriptsize (d) Learning progress}
  \par\vspace{-5pt}
  \caption{Double-integrator parameter learning from $\theta_0\in\{0.1,10\}$. Barrier smoothing rounds active-set kinks in the outer loss (a) and gradient (b). This allows gradient descent to reach  $\theta^\star=1.10$ from both initializations (c) and accelerates convergence (d).}
  \label{fig:bilevel-toy}
  \vspace{-5pt}
\end{figure}

We illustrate this change's effect on learning via a double-integrator example, where
the state $x_k=(p_k,v_k)$ is position and velocity, and input $u_k$ is acceleration. We learn a scalar tracking weight $\theta$ over horizon $H=6$ and choose an unreachable reference under the control bound to induce active-set changes. For fixed $\theta$ and initial state $x_0$, the inner problem is
\begin{equation}
\resizebox{\columnwidth}{!}{$
\label{eq:toy-inner}
\begin{aligned}
\min_{x,u}\quad
&\sum_{k=0}^{H-1}
\left(
  \tfrac{\theta}{2}Q^{\mathrm{inner}}_k(p_k-\bar p_k)^2
  +\tfrac{1}{2}R^{\mathrm{inner}}_k u_k^2
\right)
+\tfrac{\theta}{2}Q^{\mathrm{inner}}_H(p_H-\bar p_H)^2\\
\text{s.t.}\quad
&x_{k+1}
=\begin{bmatrix}1&1\\0&1\end{bmatrix}x_k
+\begin{bmatrix}0.5\\1\end{bmatrix}u_k,\qquad
-1\le u_k\le1 .
\end{aligned}
$}
\end{equation}
The outer objective evaluates the resulting plan with different tracking and control-effort weights,
\begin{equation}
\label{eq:toy-outer}
\begin{aligned}
\min_{\theta}\ L(\theta)
={}&
\sum_{k=0}^{H-1}
\left(
  \tfrac{1}{2}Q^{\mathrm{outer}}_k
  (p_k(\theta)-\bar p_k)^2
  +\tfrac{1}{2}R^{\mathrm{outer}}_k u_k(\theta)^2
\right)\\[-1pt]
&+\tfrac{1}{2}Q^{\mathrm{outer}}_H
(p_H(\theta)-\bar p_H)^2 .
\end{aligned}
\end{equation}
We choose $Q^{\mathrm{outer}}\ne Q^{\mathrm{inner}}$ and $R^{\mathrm{outer}}\ne R^{\mathrm{inner}}$, so learning $\theta$ balances tracking accuracy against control effort. We apply gradient descent with $\eta=0.1$ from $\theta_0\in\{0.1,10\}$.

Let $L_\kappa$ denote the outer loss induced by the barrier-smoothed inner problem, with $\kappa=0$ denoting the original hard NLP. Fig.~\ref{fig:bilevel-toy}(a),(b) shows that the hard problem exhibits kinks in $L_\kappa$ and jumps in its gradient at active-set changes, while increasing $\kappa$ smooths these transitions. Consequently, the hard and weakly smoothed problems stall under gradient descent, whereas sufficient smoothing crosses the active-set transitions and converges to $\theta^\star=1.1$ from both initializations  (Fig.~\ref{fig:bilevel-toy}(c)), accelerating convergence  (Fig.~\ref{fig:bilevel-toy}(d)). This example motivates log-barrier smoothing for more reliable gradient-based learning of MPC parameters.

\section{\solverName{}: Smoothed SQP--ADMM}
\label{sec:method}
We propose an efficient solver returning smooth gradients, by incorporating log-barrier smoothing into TurboMPC's SQP--ADMM solver~\cite{turbompc}. Two design choices ensure efficient solves. First, in the SQP loop, we keep the log smoothing term when formulating the convex problems, instead of formulating a quadratic approximation. Second, the inner ADMM solver uses a custom splitting with a closed-form log-barrier proximal update to handle the log smoothing term for efficient updates.  

\subsection{Smoothed SQP Formulation}
\label{sec:sqp-admm}

We incorporate log-barrier smoothing into TurboMPC's SQP--ADMM
solver~\cite{turbompc}. For a fixed barrier parameter $\kappa>0$, we consider the smoothed NLP~\cite{tracy2024ipm,frey2025diffnmpc},
\begin{equation}
\label{eq:snlp}
\begin{aligned}
\min_{x,\nu}\quad
&\ell(x;\theta)-\kappa\textstyle\sum_{i=1}^{p}\log \nu_i\\
\text{s.t.}\quad
&f(x;\theta)=0,\\
&g(x;\theta)+\nu=h(\theta),\qquad \nu>0.
\end{aligned}
\end{equation}
Stationarity with respect to $\nu$ gives $y_g=\kappa/\nu$, replacing
hard complementarity with $\nu\circ y_g=\kappa\ones$, $\nu, y_g>0$.

At each SQP iterate $x^j$, we form a quadratic model of the original
objective $\ell$ and linearize the constraints $f$ and $g$~\cite{nocedalwright},
retaining the convex log-barrier term exactly:
\begin{equation}
\label{eq:barrier-sub}
\begin{aligned}
\min_{x,\nu}\quad
&\half x^\top Px+q^\top x
-\kappa\textstyle\sum_{i=1}^{p}\log \nu_i\\
\text{s.t.}\quad
&Cx=c,\qquad \nu=\bar h-Gx>0.
\end{aligned}
\end{equation}
Here $C$ and $G$ are evaluated at $x^j$, $P\succeq0$ approximates the Lagrangian Hessian, typically Gauss--Newton~\cite{nocedalwright}. We define,
\begin{equation}
\label{eq:sqp-data}
\begin{aligned}
q&:=\nabla_x\ell(x^j;\theta)-Px^j,
&c:=Cx^j-f(x^j;\theta),\\
\bar h&:=h(\theta)-g(x^j;\theta)+Gx^j.
\end{aligned}
\end{equation}
Although the retained log barrier makes \eqref{eq:barrier-sub} nonquadratic, the subproblem remains convex, enabling efficient ADMM updates through operator splitting~\cite{boyd2011admm,osqp}.

\subsection{SAQP: Smoothed ADMM QP Solver}
\label{sec:smoothed-admm}

We solve \eqref{eq:barrier-sub} using \texttt{SAQP} (smoothed ADMM QP), which replaces the inequality projection in standard ADMM with a closed-form log-barrier proximal update (Algorithm~\ref{alg:saqp}).
\texttt{SAQP} is the inner QP solver of \solverName{}, whose outer SQP
loop solves the smoothed NLP~\eqref{eq:snlp}.
We stack the constraint matrices,
$A:=\AverageSmallMatrix{C\\G}$, and introduce $z:=(z_f,z_g)$.
The constraint $Ax=z$ gives the
equivalent split problem,
\begin{equation}
\label{eq:split-smooth}
\begin{aligned}
\min_{x,z,\nu}\quad
&\half x^\top Px+q^\top x+\mathcal I_{\{c\}}(z_f)
-\kappa\textstyle\sum_{i=1}^{p}\log \nu_i\\
\text{s.t.}\quad
&Ax-z=0,\qquad \nu=\bar h-z_g>0.
\end{aligned}
\end{equation}
Here $\mathcal I_{\{c\}}(z_f)=0$ if $z_f=c$, and $+\infty$ otherwise.
We use multipliers $y:=(y_f,y_g)$ and a block-diagonal ADMM penalty $\rho:=\blkdiag(\rho_f I_m,\rho_g I_p)$, with $\rho_f,\rho_g>0$.
Following the over-relaxed ADMM updates of~\cite{osqp,turbompc}, $\sigma>0$ regularizes the primal solve and $\alpha\in(0,2)$ controls over-relaxation.
We use the convergence check and adaptive $\rho$ update from~\cite{turbompc}.

\textbf{SAQP updates:}\quad
\texttt{SAQP} modifies standard ADMM's slack update, replacing the projection with a closed-form update resulting from applying the proximal operator to the log-barrier term~\cite{boyd2011admm,osqp}. One ADMM iteration is\\
\begin{subequations}
\label{eq:admm}
\vspace{-5pt}
\begin{tcolorbox}[
  enhanced,colback=red!3,colframe=red!9,boxrule=0.25pt,arc=1pt,
  boxsep=0pt,left=0pt,right=0pt,top=0pt,bottom=0pt,
  before skip=0pt,after skip=1pt]
\noindent{\footnotesize\textbf{Primal update}}\par\vspace{-10pt}
{\small
\begin{equation}
\label{eq:admm-primal}
\begin{aligned}
S&:=P+\sigma I+A^\top\rho A\succ0,\\
S\,\tilde x^{k+1}
&=\sigma x^k-q+A^\top\!\bigl(\rho z^k-y^k\bigr),
\quad \tilde z^{k+1}=A\tilde x^{k+1},\\
(\hat x,\hat z)^{k+1}
&=\alpha(\tilde x,\tilde z)^{k+1}+(1{-}\alpha)(x,z)^k,
\quad x^{k+1}=\hat x^{k+1}.
\end{aligned}
\end{equation}
}
\end{tcolorbox}
\begin{tcolorbox}[
  enhanced,colback=ForestGreen!3,colframe=ForestGreen!9,boxrule=0.25pt,arc=1pt,
  boxsep=0pt,left=0pt,right=0pt,top=0pt,bottom=0pt,
  before skip=0pt,after skip=1pt]
\noindent{\footnotesize\textbf{Log-barrier slack update}}\par\vspace{-7pt}
{\small
\begin{equation}
\label{eq:retr-update}
\begin{alignedat}{2}
\bar z_g^{k+1} &:=\hat z_g^{k+1}+\rho_g^{-1}y_g^k,
&\quad r^{k+1} &:=\bar h-\bar z_g^{k+1},\\
\nu^{k+1} &=b_\Gamma\bigl(r^{k+1}\bigr),
&\quad \Gamma &:=\kappa\rho_g^{-1},\\
z_g^{k+1} &=\bar h-\nu^{k+1},
&\quad z_f^{k+1} &=c.
\end{alignedat}
\end{equation}
}
\end{tcolorbox}
\begin{tcolorbox}[
  enhanced,colback=blue!3,colframe=blue!9,boxrule=0.25pt,arc=1pt,
  boxsep=0pt,left=0pt,right=0pt,top=0pt,bottom=0pt,
  before skip=0pt,after skip=1pt]
\noindent{\footnotesize\textbf{Dual update}}\par\vspace{-20pt}
\begin{equation}
\label{eq:admm-dual}
y^{k+1}=y^k+\rho\bigl(\hat z^{k+1}-z^{k+1}\bigr).
\end{equation}
\end{tcolorbox}
\end{subequations}

\eqref{eq:retr-update} solves the proximal problem
componentwise:
\begin{equation}
\label{eq:zprox}
z_g^{k+1}=\argmin_{z<\bar h}
-\kappa\log(\bar h-z)
+\tfrac{\rho_g}{2}(z-\bar z_g^{k+1})^2.
\end{equation}
With $\nu=\bar h-z$, stationarity gives
$\nu^2-r^{k+1}\nu-\Gamma=0$. Its positive root is
\vspace{-7pt}
\begin{equation}
\label{eq:b}
b_\Gamma(v):=\frac{v+\sqrt{v^2+4\Gamma}}{2},\qquad \Gamma>0.
\end{equation}

\begin{remark}[Per-iteration smoothed complementarity]
\label{rmk:complementarity}
For $\Gamma>0$, $b_\Gamma(v)>0$ and $b_\Gamma'(v)\in(0,1)$, so the slack update is smooth and yields $\nu^{k+1}>0$. Together with the dual update, it gives
\begin{equation}
\label{eq:headline}
\boxed{\,
y_g^{k+1}=\frac{\kappa}{\nu^{k+1}}
\quad\Longrightarrow\quad
\nu^{k+1}\circ y_g^{k+1}=\kappa\ones.
\,}
\end{equation}
Smoothed complementarity holds after every ADMM iteration within a \texttt{SAQP} solve for the split slack $\nu^{k+1}=\bar h-z_g^{k+1}$.
\end{remark}

\begin{remark}[Recovery of original NLP and QP]
\label{rmk:recovery}
As $\kappa\to0^+$, smoothed complementarity recovers
$\nu\circ y_g=0$, and any finite limit of smoothed KKT solutions
satisfies the original NLP-KKT conditions~\cite{tracy2024ipm,frey2025diffnmpc}.
The smoothed subproblem recovers
\begin{equation}
\label{eq:qp-limit}
\min_x\ \half x^\top Px+q^\top x
\quad\text{s.t.}\quad Cx=c,\quad Gx\leq\bar h.
\end{equation}
For fixed $\rho_g>0$, $b_\Gamma(v)\to\max(v,0)$, so the slack update
recovers the standard ADMM projection~\cite{osqp}.
\end{remark}

\begin{remark}[Two-sided constraints]
\label{rmk:two-sided}
MPC constraints are often two-sided,
$h_l(\theta)\le g(x;\theta)\le h_u(\theta)$, e.g., representing input and track bounds.
Instead of stacking them as $\AverageSmallMatrix{G\\-G}$, which doubles the number of 
inequality rows, \solverName{} applies the barrier
$-\kappa\log\nu_l-\kappa\log\nu_u$ per row, with
$\nu_l:=z_g-\bar h_l$ and $\nu_u:=\bar h_u-z_g$.
The slack update \eqref{eq:zprox} then becomes the unique root of a strictly
monotone scalar equation. We compute it elementwise with a fixed number of
safeguarded Newton steps, warm-started from $b_\Gamma$, to machine precision
(Appendix~\ref{app:two-sided}).
Remarks~\ref{rmk:complementarity}--\ref{rmk:recovery} hold per bound, and
$\kappa\to0^+$ recovers the box projection of~\cite{osqp}.
\end{remark}

\subsection{Implicit Differentiation and GPU Computation}
\label{sec:implicit-gpu}

For the backward pass (Algorithm~\ref{alg:vjp}), we use the exact NLP Lagrangian Hessian $P=\mathcal H$ the KKT system of the smoothed subproblem~\eqref{eq:barrier-sub}:\begin{equation}
\label{eq:pkkt}
\begin{gathered}
F^{\mathrm{QP}}_{\kappa}(w;\theta):=
\begin{bmatrix}
Px+q+C^\top y_f+G^\top y_g\\
Cx-c\\
\nu\circ y_g-\kappa\ones
\end{bmatrix}=0,\\
\nu:=\bar h-Gx>0,\qquad y_g>0.
\end{gathered}
\end{equation}
At SQP convergence, \eqref{eq:sqp-data} makes this residual
equal to the smoothed NLP-KKT residual:
\begin{equation}
\label{eq:consistency}
F^{\mathrm{QP}}_{\kappa}(w^\star;\theta)
=F^{\mathrm{NLP}}_{\kappa}(w^\star;\theta)=0.
\end{equation}
Assume twice continuously differentiable problem data, linear independence
constraint qualification (LICQ), and second-order sufficient conditions
(SOSC) for the smoothed problem at this solution.
Together with $\nu\circ y_g=\kappa\ones$ and $\nu,y_g>0$, these
conditions make the KKT Jacobian nonsingular~\cite{frey2025diffnmpc}.
With the exact Hessian, the QP and NLP KKT linearizations coincide at convergence, yielding matching sensitivities of a locally $C^1$ solution map.

We exploit stagewise sparsity and batch parallelism to compute both
solver passes efficiently on GPUs.

\textbf{Forward pass.}
For fixed QP data and $\rho$, the stage-sparse matrix $S$ in
\eqref{eq:admm-primal} is constant, enabling reuse of its factorization across iterations.
The elementwise slack update preserves the per-iteration complexity of
standard ADMM.

\textbf{Backward pass.}
Eliminating $\lambda_g=-D^{-1}G\lambda_x$, with $D:=\diag(\nu)$,
from \eqref{eq:adjoint} yields
\begin{equation}
\label{eq:folded}
\begin{bmatrix}
\mathcal H+G^\top W^{-1}G & C^\top\\
C & 0
\end{bmatrix}
\begin{bmatrix}\lambda_x\\\lambda_f\end{bmatrix}
=-\begin{bmatrix}\nabla_xL\\0\end{bmatrix},
\end{equation}
where $W^{-1}:=\diag(y_g/\nu)=\diag(\kappa/\nu^2)$
provides continuously varying inequality weights.
With $\bar\lambda_g:=W^{-1}G\lambda_x$, the parameter VJP is
\begin{equation}
\label{eq:grads}
\nabla_\theta L
=[\partial_\theta\nabla_x\mathcal L]^\top\lambda_x
+[\partial_\theta f]^\top\lambda_f
+[\partial_\theta(g-h)]^\top\bar\lambda_g.
\end{equation}
For cost-only parameters, the last two terms vanish.

Both systems retain the MPC problem's stagewise sparsity. 
We refer readers to~\cite{schwan2025multistage,jordana2025structure,turbompc} for numerical methods that exploit this structure.
\solverName{} reuses TurboMPC's structured GPU kernels and solves independent problems in parallel across the batch~\cite{turbompc}.
\vspace{-5pt}
\begin{remark}[MPC feasibility during training]
\label{rmk:feasibility}
To prevent inequality constraints from making MPC infeasible as the policy
changes during training, we relax them to $g(x;\theta)\leq h(\theta)+\xi$
and add the penalty $\gamma_\xi\|\xi\|^2/2$, $\gamma_\xi>0$, to the
objective~\cite{admmslack}. The slack update \eqref{eq:retr-update} remains
closed form with $\Gamma=\kappa(\rho_g^{-1}+\gamma_\xi^{-1})$,
$\xi^{k+1}=\kappa/(\gamma_\xi\nu^{k+1})$, and
$z_g^{k+1}=\bar h-\nu^{k+1}+\xi^{k+1}$.
The backward weights \eqref{eq:folded} become $W^{-1}=\diag(\tfrac{y_g}{\nu+\gamma_\xi^{-1}y_g}).$

\end{remark}

\section{Efficient Reinforcement Learning with \solverName{}}
\label{sec:rl}
Next, we co-design the RL framework to enable efficient training of differentiable MPC policies.

\subsection{MPC Policies for Reinforcement Learning}
\label{sec:rl-problem}

MPC policies incorporate model knowledge and explicit constraints through an optimal control problem (OCP)~\cite{turbompc}:
\begin{equation}
\label{eq:ocp}
\begin{aligned}
\min_{x,u}\quad
&\textstyle\sum_{t=0}^{H-1}\ell_t(x_t,u_t,x_{t+1},u_{t+1};\theta)+\ell_H(x_H;\theta)\\
\text{s.t.}\quad
&x_0=x_{\mathrm{init}},\\
&f_t(x_t,u_t,x_{t+1},u_{t+1};\theta)=0,\quad t=0,\ldots,H{-}1,\\
&\underline g_t\leq g_t(x_t,u_t;\theta)\leq\overline g_t,\quad t=0,\ldots,H.
\end{aligned}
\end{equation}
Here $x,u$ are predicted states and controls, $x_{\mathrm{init}}$ is the
initial state, $H$ is the horizon, and $\ell,f,g$ specify costs, dynamics, and constraints. The policy applies the first control $u_0^\star(\theta,x_{\mathrm{init}})$ and then resolves~\eqref{eq:ocp} for the new $x_{\mathrm{init}}$ at the next control step.

Given environment state $s$, a neural network with weights $\phi$ predicts
state-dependent MPC parameters $\theta(s)=\pi_\phi^\theta(s)$.
For exploration, we use a Gaussian policy~\cite{schulman2017ppo}:
\begin{equation}
\label{eq:policy}
\pi_\phi(a\mid s)
=
\mathcal N\!\left(
a;\,
u_0^\star(\theta(s),x_{\mathrm{init}}),
\Sigma
\right),
\end{equation}
where $\Sigma$ is the exploration covariance. RL optimizes $\phi$ to
maximize $J(\phi)=\E_{\pi_\phi}[\sum_{t=0}^{\infty}\gamma^t r_t]$,
with reward $r_t$ and discount $\gamma\in[0,1)$.
Differentiating the MPC solution map provides policy sensitivities to the learned parameters and initial state, enabling gradient propagation through both the parameter network and closed-loop rollouts.
\vspace{-4pt}
\begin{figure}[!t]
\centering
\begin{tcolorbox}[
diffmpcOuterFull,
title={
\texttt{\solverName{}} (Differentiable MPC Solver)
\phantom{${}^1$}}
]

\begin{tcolorbox}[
diffmpcADMMFull,
title={
\textbf{Alg. \customlabel{alg:saqp}{1}}\quad
\texttt{SAQP} (Smoothed QP Solver)
\phantom{${}^1$}
}
]
{\small\noindent\textcolor{blue}{// Inner solve at each SQP iteration of the forward pass.}\par}
\begin{algorithmic}[1]
\Statex \hspace{-5mm}\textbf{Inputs}:
$P,q,A,c,\bar h,\sigma,\rho_f,\rho_g,\alpha,\kappa$
\Statex \hspace{-5mm}\textbf{Initial guess}: $x^0,z^0,y^0$

\While{not converged}
  \State \textbf{Primal update}
  \newalgcomment{Eq.}{eq:admm-primal}

  \State \textbf{Log-barrier slack update}
  \newalgcomment{Eq.}{eq:retr-update}

\State \textbf{Dual update} {\footnotesize\eqref{eq:admm-dual}}  {\scriptsize\Comment{Smoothed complementarity holds}}

  \State Check convergence; Update $\rho$~\cite{turbompc}
\EndWhile

\State \textbf{Return}: forward iterate $\hat w$
\end{algorithmic}
\end{tcolorbox}

\begin{tcolorbox}[
diffmpcBackwardFull,
title={
\textbf{Alg. \customlabel{alg:vjp}{2}}
Smoothed \texttt{VJP} (Backward Pass)
\phantom{${}^1$}
}
]
\begin{algorithmic}[1]
\Statex \hspace{-5mm}\textbf{Inputs}: $\hat w$, $\theta$, $\nabla_xL$

\State Form exact-Hessian smoothed KKT linearization

\State Solve reduced adjoint system for $(\lambda_x,\lambda_f)$
\newalgcomment{Eq.}{eq:folded}

\State $\bar\lambda_g\gets W^{-1}G\lambda_x$

\State Compute parameter \texttt{VJP}
\newalgcomment{Eq.}{eq:grads}

\State \textbf{Return}: gradient $\nabla_\theta L$
\end{algorithmic}
\end{tcolorbox}

\end{tcolorbox}
\end{figure}

\begin{figure}[!t]
\centering
\begin{tcolorbox}[
diffmpcTrainFull,
title={
\textbf{Alg. \customlabel{alg:shac}{3}}
\texttt{FORL} with \texttt{\solverName{}}
\phantom{${}^1$}
}
]
\begin{algorithmic}[1]
\Statex \hspace{-5mm}\textbf{Inputs}:
$\pi_\phi^\theta$, $\Sigma$, $V_\psi$, $V_{\bar\psi}$, $\mathrm{env}$, $T$
\Statex \hspace{-5mm}\textbf{Initialize}:
$s_0\gets\mathrm{reset}(B)$; seed warm plans

\While{training}

  \For{$t=0,\dots,T-1$}
    \State $\theta_t\gets\pi_\phi^\theta(s_t)$

     \State $(x^\star,u^\star,y^\star)\gets\mathrm{SQP}$
     \newalgcomment{Inner: Alg.}{alg:saqp}
    \State $a_t\sim\mathcal N(u_0^\star,\Sigma)$
    \newalgcomment{Eq.}{eq:policy}

    \State $(s_{t+1},r_t)\gets\mathrm{env}(s_t,a_t)$

    \State Accumulate bootstrapped $J(\phi)$
    \newalgcomment{Eq.}{eq:shac-objective}

    \State Shift warm plans; reset finished rows
  \EndFor

  \State \textbf{Actor update}: backpropagate through environment and
  the MPC solution map
  \newalgcomment{Backward: Alg.}{alg:vjp}

  \State \textbf{Critic update}: fit TD($\lambda$), update
  $V_{\bar\psi}$
  \newalgcomment{Eq.}{eq:tdlambda}

  \State $s_0\gets s_T$
\EndWhile

\State \textbf{Return}: trained $\phi$
\end{algorithmic}
\end{tcolorbox}
\vspace{-10pt}
\end{figure}

\subsection{The Computational Cost of Learning MPC Policies}
\label{sec:why-forl}
\vspace{-2pt}
For an MPC policy, each action evaluation requires a forward optimization
solve, while differentiating that action requires a backward pass.
Consequently, reusing collected states after the policy changes requires
additional MPC solves.

We compare three learning strategies:
FORL, which differentiates a rollout for one policy update~\cite{xu2022shac}; 
PPO, which reuses on-policy data over multiple optimization epochs~\cite{schulman2017ppo,romero2024acmpc}; 
and off-policy actor--critic methods, which repeatedly sample replayed states for critic and actor updates (e.g., TD3, SAC)~\cite{fujimoto2018td3,haarnoja2018sac,chen2021redq}.

We approximate the computation of one learning iteration:
\begin{equation}
\label{eq:update-cost}
\mathcal C_{\mathrm{update}}
\simeq
n_{\mathrm{env}}\mathcal C_{\mathrm{env}}
+n_{\mathrm{fwd}}\mathcal C_{\mathrm{fwd}}
+n_{\mathrm{bwd}}\mathcal C_{\mathrm{bwd}},
\end{equation}
where $\mathcal C_{\mathrm{env}}$, $\mathcal C_{\mathrm{fwd}}$, and
$\mathcal C_{\mathrm{bwd}}$ are the average costs of an environment step, MPC forward solve, and MPC backward pass.
The subscripted $n$ terms count each operation.

\begin{figure}[!t]
\centering
\begingroup
\makeatletter
\def\@captype{table}
\makeatother
\caption{Profiled MPC solver work per learning iteration.}
\vspace{-5pt}
\label{tab:solves}
\centering
\footnotesize
\setlength{\tabcolsep}{2.5pt}
\begin{tabular}{@{}lcccc@{}}
\toprule
& \multicolumn{2}{c}{Forward} & \multicolumn{2}{c}{Backward}\\
\cmidrule(lr){2-3}\cmidrule(l){4-5}
Algorithm & count & time (s) & count & time (s)\\
\midrule
MPC + FORL (ours)
& $N$ & 0.63
& $N$ & 0.09\\
MPC + PPO~\cite{romero2024acmpc}
& $(1{+}E)N$ & 7.39
& $EN$ & 0.96\\
MPC + TD3
& $(1{+}nM{+}\tfrac{nM}{d})N$ & 22.16
& $\tfrac{nM}{d}N$ & 0.95\\
\bottomrule
\end{tabular}
\par\smallskip
\begin{minipage}{\columnwidth}
\footnotesize\raggedright
$N=BT$ transitions; profile uses $B=1{,}024$, $T=8$, $E=10$,
$M=1{,}024$, $n=1/50$, and $d=2$. Here, $M$ is the replay minibatch size, and the actor is updated every $d$ critic updates. The conservative update-to-data
ratio $n$ understates the off-policy gap.
\end{minipage}
\par
\endgroup

\vspace{0pt}
\includegraphics[width=\columnwidth]{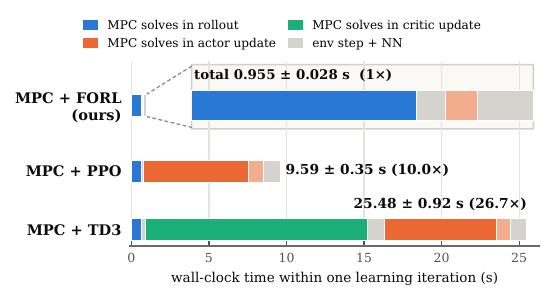}
\par\vspace{-12pt}
\caption{Measured wall-clock breakdown of one learning update
(mean over five trials). Lighter blocks denote backward passes. FORL
update is magnified.}
\label{fig:update-structure}
\vspace{-5pt}
\end{figure}

Table~\ref{tab:solves} reports MPC solver-pass counts for the racing setup in Sec.~\ref{sec:exp-training}. FORL differentiates retained rollout solutions
  once per update; PPO repeats forward and backward passes over $E$ epochs,
  while TD3/SAC re-evaluates MPC for replayed critic targets and actor updates.
  Profiled PPO and TD3 updates take approximately $10\times$ and $27\times$
  the wall-clock time of FORL, respectively (Fig.~\ref{fig:update-structure}).
  The expensive repeated MPC solves imply that data reuse can reduce environment
  interactions while increasing training time. This reverses the usual
  balance for direct neural policies, where environment interaction often
  dominates computation~\cite{gleeson2021rlscope}.

\subsection{Computationally Efficient First-Order RL with MPC}
\label{sec:training}

These observations motivate our choice of FORL, instantiated as
Short-Horizon Actor-Critic (SHAC)~\cite{xu2022shac}. Its actor is the MPC
policy, and its critic $V_\psi(s)$ learns the expected discounted return
from state $s$ under that policy. The target critic $V_{\bar\psi}$ is a
slowly updated copy of $V_\psi$, used to stabilize both policy updates and
critic training. SHAC differentiates short-rollout rewards plus the target
critic's estimate of the remaining return, avoiding full episode backpropagation. 
We get environment derivatives from a differentiable simulator and note that learned
dynamics can provide surrogate gradients~\cite{amigo2025decoupled}.

\textbf{Actor update.}
Each update collects a differentiable rollout of $T$ steps in $B$ parallel
environments.
For a rollout without resets, the actor maximizes
\begin{equation}
\label{eq:shac-objective}
J(\phi)
=
\frac{1}{BT}
\sum_{b=1}^{B}
\left[
\sum_{t=0}^{T-1}\gamma^t r_t^b
+\gamma^T V_{\bar\psi}(s_T^b)
\right].
\end{equation}
We use reparameterized Gaussian actions and update $\phi$ once per rollout.
Gradients flow through the environment, the MPC solution map, and the
target critic's state input while its weights remain fixed.
We retain the MPC solutions and associated backward-pass data during rollouts, allowing the actor update to compute implicit sensitivities without additional MPC forward solves.
States are detached between rollouts.

\textbf{Critic update.}
Using the same rollout, we fit $V_\psi$ to TD($\lambda$)
targets~\cite{sutton1988td}, with $\hat V_T=V_{\bar\psi}(s_T)$  :
\begin{equation}
\label{eq:tdlambda}
\hat V_t
=
r_t+\gamma
\left[
(1-\lambda)V_{\bar\psi}(s_{t+1})
+\lambda\hat V_{t+1}
\right].
\end{equation}
We bootstrap at rollout truncations, use zero continuation at true
terminals, and update the target critic by exponential moving average (EMA).
Critic fitting uses detached states and targets, and requires no additional MPC solves.
Algorithm~\ref{alg:shac} summarizes the pipeline.

\begin{figure*}[!t]
\vspace{-10pt}
  \centering
  \includegraphics[width=\textwidth]{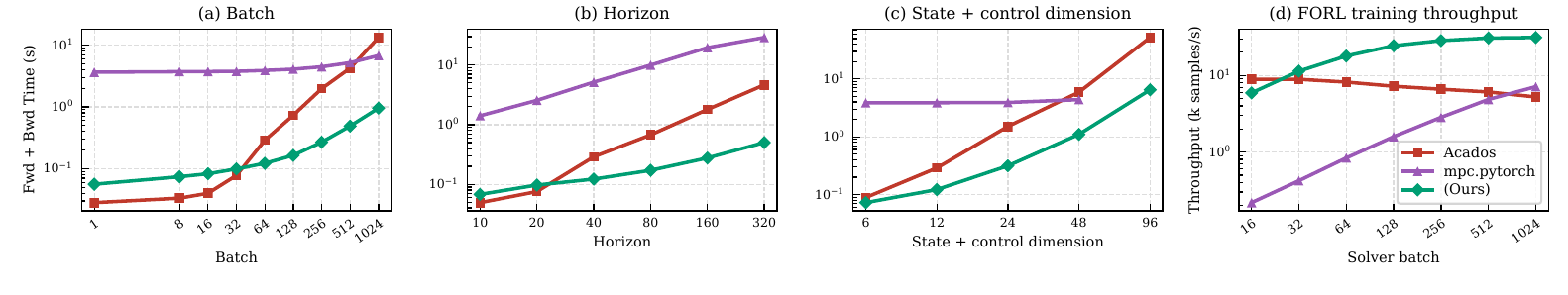}
  \par\vspace{-18pt}
\caption{Constrained linear-system comparison of \solverName{}, acados,
  and mpc.pytorch: (a)--(c) forward-plus-backward time versus batch size $B$,
  horizon $H$, and state--control dimension $n_x+n_u$, respectively;
  (d) end-to-end SHAC throughput versus solver batch size $M$ ($B=1024$).}
  \label{fig:solver-efficiency}
  \vspace{-4pt}
\end{figure*}

\section{Experiments}
\label{sec:experiments}

We perform a three part evaluation of our approach. First we demonstrate the scalability and throughput of our solver on benchmark tasks.
Then, we demonstrate efficient learning of MPC policies on a simulated autonomous racing benchmark. 
Finally, to assess policy robustness, we train with domain randomization and deploy the policy on a Lexus LC500.

\vspace{-5pt}
\subsection{Speed and Scalability}
\label{sec:exp-solver-efficiency}

We evaluate on the constrained linear-system benchmark of
TurboMPC~\cite{turbompc} using a single NVIDIA
RTX~5090 GPU and an Intel Core Ultra~9~285K CPU. The nominal settings are state and control dimensions $(n_x,n_u)=(8,4)$,
MPC horizon $H=40$,
rollout horizon $T=50$, control bound $u_{\max}=10$, and solver tolerance
$10^{-3}$. The scaling tests use problem batch size $B=64$ unless varied, while
training uses $B=1024$.

\textbf{1) Solver scaling:}\;
Figure~\ref{fig:solver-efficiency}(a)--(c) compares \solverName{}~
with acados~\cite{verschueren2022acados} on the CPU and
mpc.pytorch~\cite{amos2018diffmpc} on the GPU, measuring the combined time
for a 50-step closed-loop rollout and differentiation of its cumulative
reward. At problem batch size $B=1$, acados is $2.0\times$ faster than
\solverName{}. As the problem batch grows to $B=1024$
(Fig.~\ref{fig:solver-efficiency}(a)), \solverName{}~ achieves speedups of
$13.9\times$ over acados and $7.1\times$ over mpc.pytorch. Batched
execution therefore benefits such parallel policy training workloads.

The horizon sweep in Fig.~\ref{fig:solver-efficiency}(b) shows similar
scaling benefits: at $H=320$, \solverName{}~achieves speedups of $9.3\times$
over acados and $57.8\times$ over mpc.pytorch. At the largest
state--control dimension, $n_x+n_u=96$
(Fig.~\ref{fig:solver-efficiency}(c)), \solverName{}~is $8.1\times$ faster
than acados, while mpc.pytorch runs out of GPU memory.
We also observe that \solverName{}~matches TurboMPC's computational performance.

\textbf{2) Training throughput:}\;
Figure~\ref{fig:solver-efficiency}(d) reports end-to-end throughput for
SHAC training of the actor and critic neural networks at fixed problem
batch size $B=1024$, varying the solver batch size
$M\in\{16,32,64,128,256,512,1024\}$, the number of OCPs processed
together. Our throughput increases with $M$,
with the highest tested throughput at $M=1024$. The best tested solver
batch sizes for acados and mpc.pytorch are $M=16$ and $M=1024$,
respectively. Using the best batch size for each method,
our pipeline achieves $3.5\times$ the throughput of acados and
$4.4\times$ of mpc.pytorch. Thus our scaling advantage
carries through to full learning updates.

\textbf{Efficient MPC policy learning depends on both solver speed and the
number of solves required by the learning algorithm.}
Reusing data for additional policy updates can reduce sample requirements
while increasing costly MPC solves. Our approach addresses both:
\solverName{} accelerates batched solves, while FORL limits repeated MPC
evaluations during policy updates (Fig.~\ref{fig:update-structure}).

\vspace{-3pt}
\subsection{Efficient RL of MPC Policies}
\vspace{-2pt}
\label{sec:exp-training}
Next, we consider the challenging problem of minimum-time racing of a vehicle with stiff nonlinear dynamics. The problem requires driving at the track bounds and actuator limits, with frequent changes in the active constraints that motivate using log-smoothed constraints. We train the policies in simulation using the same computer as Section~\ref{sec:exp-solver-efficiency}.

\textbf{1) Racing setup:}\;
We first compare learning algorithms on a closed-oval trajectory-tracking task. The MPC problem is a single-track spatial-domain OCP \cite[Section V]{turbompc}. The state $x=(r,v,\beta,\omega_r,e,\Delta\phi,t)$ comprises
yaw rate, speed, sideslip, rear-wheel speed, lateral deviation, relative
heading, and elapsed time. 
The control input $u=(\delta,\tau_r,\tau_f)$ comprises
steering angle, signed rear drive/brake torque, and front braking torque. 
We use a quadratic running cost
\begin{equation}
\label{eq:racing:cost}
\ell_t\!(x_t,u_t,u_{t+1})
\!=\!\|x_t\!-\!x_{t,\mathrm{ref}}\|_{Q_\theta}^{2}
\!+\!\|u_t\!-\!u_{t,\mathrm{ref}}\|_{R_\theta}^{2}
\!+\!\|u_t\!-\!u_{t+1}\|_{W}^{2}.
\nonumber
\end{equation}
where $(Q_\theta,R_\theta,W)$ are diagonal matrices.
The terminal cost is $\ell_H(x_H)=\|x_H-x_{H,\textrm{ref}}\|_{Q^H_\theta}^2+w_t t_H$. The weights $(W,w_t)$ are fixed and $(Q_\theta,R_\theta, Q^H_\theta)$ are output by a neural network.

To limit model mismatch when comparing learning algorithms, we use a
simulator that omits load transfer and shares the MPC's nominal vehicle
parameters, with tire-friction
coefficients $\mu_f=1.02$ (front) and $\mu_r=1.08$ (rear). The simulator
uses fixed-step Tsit5 integration in Diffrax~\cite{kidger2021on}, whereas
the MPC uses implicit trapezoidal integration.

The reward penalizes weighted squared tracking errors in
$(r,v,\beta,\omega_r,e,\Delta\phi)$, with soft penalties for low speed,
track departure, and spinout, plus a penalty for early termination.
We sample initial reference states uniformly along the track and add small
perturbations. Episodes last up to $1{,}000$ control steps ($20$\,s) and
terminate early if $v<9.5$\,m/s, $|\Delta\phi|>60^\circ$, the vehicle remains off
track for more than five simulation steps, or a transition becomes nonfinite.

\begin{figure}[!t]
  \centering
  \includegraphics[width=\columnwidth]{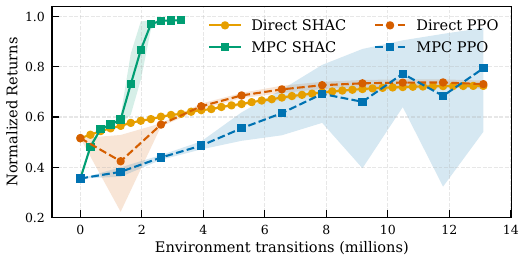}\\
  \includegraphics[width=0.492\columnwidth]{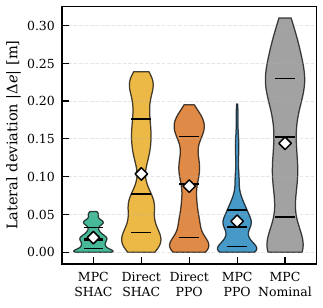}\hfill
  \includegraphics[width=0.492\columnwidth]{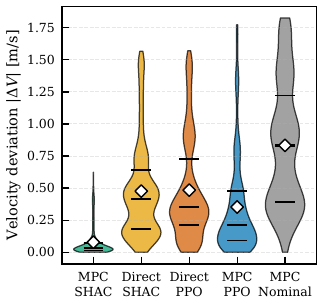}
  \par\vspace{-4pt}
  \caption{Top: normalized return versus environment transitions
  (mean and 95\% confidence interval across five seeds).
  Bottom: per-step absolute lateral and speed tracking errors,
  $|\Delta e|=|e-e_{\mathrm{ref}}|$ and
  $|\Delta v|=|v-v_{\mathrm{ref}}|$.
  Diamonds indicate means and horizontal bars indicate quartiles.}
  \label{fig:minimal-training}
  \vspace{-5pt}
\end{figure}

\textbf{2) Policies and training:}\;
The MPC policy uses a $44$-dimensional observation of scaled vehicle
states, lap-position encoding, a curvature preview covering $96$\,m ahead
along the track, and last applied controls. Its cost network uses two fully connected hidden layers of 256 units with ReLU activations and a 17-dimensional linear output layer (81,681 parameters). We apply a
sigmoid to the network's raw output logits and use the resulting values to
interpolate between lower and upper bounds
$\theta_{\min}$ and $\theta_{\max}$, ensuring
$Q_{\theta},Q_{\theta}^H\succeq0$ and $R_{\theta}\succ0$. 
The network's high-dimensional parameter space would make applying Bayesian optimization techniques  \cite{turbompc} challenging, motivating the use of RL and differentiable MPC.

For a fair comparison, direct neural policy baselines add bounded steering and
drive/brake torque corrections to fixed PD lateral and PID longitudinal
control, so both policy classes improve an existing tracking controller.
We compare five methods to assess the benefits of our approach:
\begin{itemize}[leftmargin=3mm]
\setlength{\itemsep}{0pt}
\setlength{\parsep}{0pt}
\item \textbf{MPC SHAC (ours)} trains the neural MPC cost map with Alg.~\ref{alg:shac}. \item \textbf{MPC PPO (algorithm ablation)} replaces SHAC with
PPO \cite{schulman2017ppo,romero2024acmpc}, using the same neural cost map
and MPC solver.
\item \textbf{Direct SHAC (policy ablation)} trains the direct neural
policy with SHAC (Alg.~\ref{alg:shac}), replacing the MPC policy.
\item \textbf{Direct PPO (baseline)} trains the direct neural 
policy with PPO.
\item \textbf{MPC Nominal (baseline)} uses the same MPC solver with manually-tuned fixed weights as a cost adaptation baseline.
\end{itemize}

We adapt the PPO implementation from CleanRL~\cite{huang2022cleanrl}. Update horizons are $32$ control steps for SHAC and $128$ for PPO. We select each method's learning rate by grid search.

\textbf{3) Training efficiency:}\;
With the same neural cost map and MPC solver, MPC SHAC completes training
in \textbf{7\,min} using
\textbf{3.3\,million transitions}, showing the efficiency and scalability of the GPU-accelerated solver. In contrast, MPC PPO takes 5.35\,h
to train using 13\,million transitions. MPC SHAC is thus approximately
\textbf{46$\times$ faster} and uses \textbf{4$\times$ fewer samples}. 
Direct SHAC and Direct PPO each also use \textbf{13\,million} transitions and take
\textbf{5.7\,min} and \textbf{2.7\,min}, respectively.
Training times are averaged across five seeds and exclude evaluation.
These results show that combining fast MPC solves with FORL updates that limit repeated solves
(Fig.~\ref{fig:update-structure}) brings total MPC training time to the
same order of magnitude as direct neural policy training, giving significant speedups over MPC PPO.

\textbf{4) Final policy performance:}\;
MPC SHAC achieves a final normalized return of $\approx$
\textbf{0.98}, versus $0.79$ for MPC PPO and $0.72$ for both direct
policies (Fig.~\ref{fig:minimal-training}). The scale assigns zero to the
worst observed return and one to zero tracking error.
MPC SHAC also achieves the lowest mean and upper-quartile absolute
lateral and speed tracking errors, outperforming both the learned
baselines and MPC with manually-tuned fixed weights.
Thus, MPC structure improves performance over direct policies, while learned
costs improve tracking over MPC with manually-tuned fixed weights.

\textbf{5) Insights on enabling stable training of MPC policies:}\;
We observed that backpropagation through long rollouts, 
as in DiffMPC~\cite{adabag2026diffmpc}, did not lead to stable training for this racing task. Instead, 
SHAC learns successfully with short rollout backpropagation
(Fig.~\ref{fig:minimal-training}), as shorter gradient chains limit instability,
while a terminal critic accounts for rewards beyond the
rollout~\cite{xu2022shac}.
Using SHAC could also improve policy performance for TurboMPC~\cite{turbompc}. However, as TurboMPC does not include complementarity smoothing, and batched rollouts, different environments and time steps sit on different active sets, the underlying hard NLP's sensitivities will jump whenever a constraint enters or leaves the active set (Fig.~\ref{fig:bilevel-toy}). This will result in noisy aggregated policy gradients~\cite{jahncke2026diffwmpc}, which can negatively impact training stability, reinforcing the benefits of \solverName{}'s Log-barrier smoothing.

\compactsubsection{Robustness and Real-World Deployment}
\label{sec:exp-robustness}
\label{sec:exp-realworld}

\begin{figure}[!t]
  \centering
  \includegraphics[width=\columnwidth]{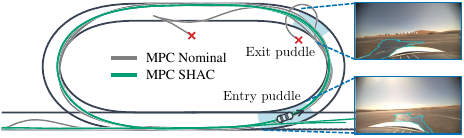}
  \includegraphics[width=\columnwidth]{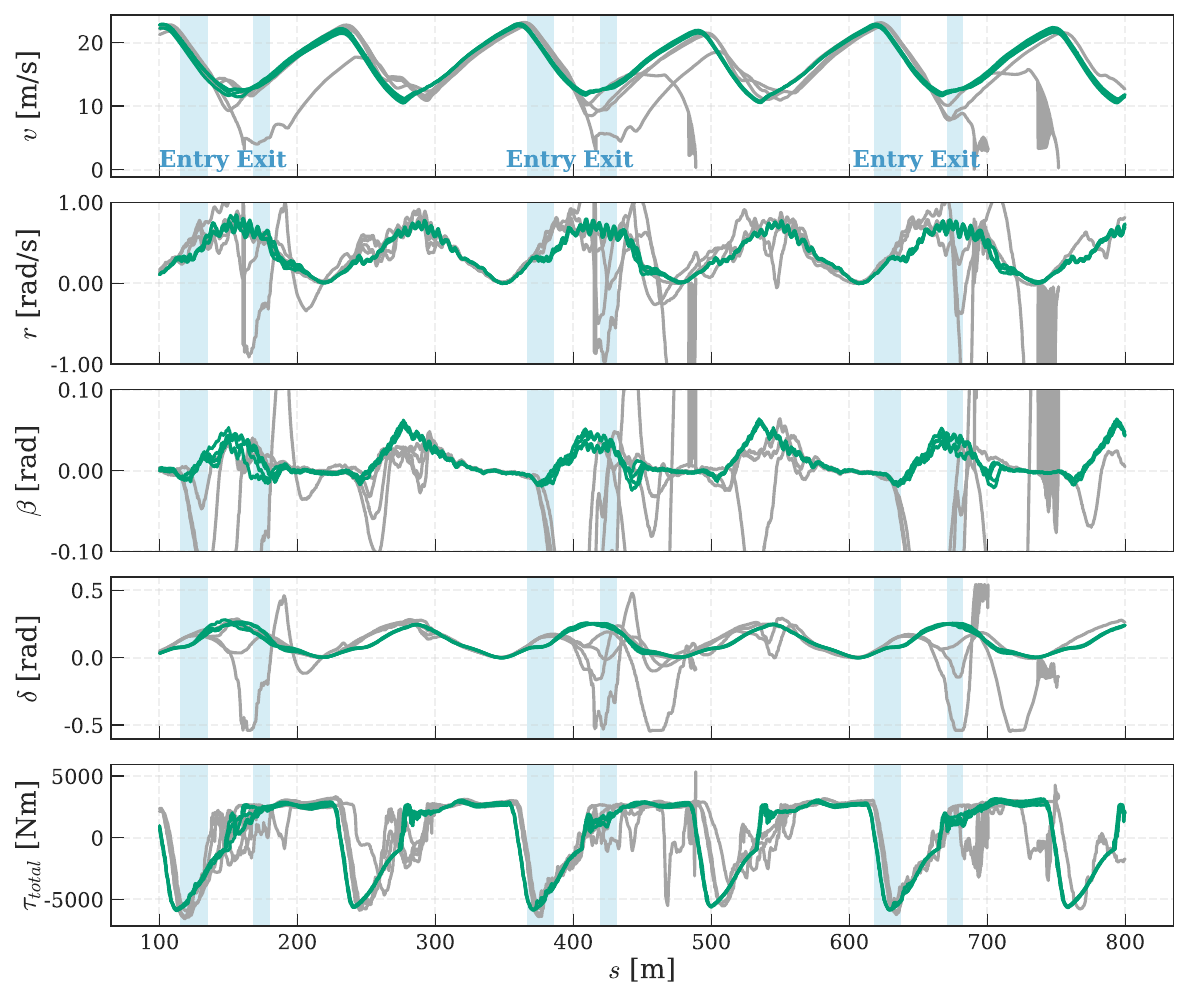}
  \par\vspace{-10pt}
  \caption{Real-world racing results. Top: one trial per controller per
  scenario, with puddles shaded blue and nominal trajectory endpoints
  marked by red crosses. Bottom: state and control traces for all trials.}
  \label{fig:realworld}
  \vspace{-5pt}
\end{figure}

\textbf{1) Training for robust deployment:}\;
We use domain randomization to improve policy robustness to unmodeled
conditions and model mismatch for real-world deployment. Unlike learning from
open-loop MPC prediction losses~\cite{jahncke2026diffwmpc}, RL with
domain randomization directly optimizes closed-loop return.
In particular, we train with randomized dry friction and localized puddles while
retaining nominal MPC dynamics.
We model load transfer $\Delta F_z$ in both the simulator and MPC and train
with a $10\%$ faster reference, $128$-step SHAC rollouts, and $30$\,s episodes.
The neural network also outputs linear state cost coefficients $q_\theta$ and $q_\theta^H$, parameterized as $q=-2Q\,\delta x_{\mathrm{ref}}$, using interpretable
bounded target offsets to reduce sensitivity differences between quadratic
and linear terms.
Also, we add path progress to the reward to promote faster racing, and the
stall threshold is set to $v <5$\,m/s.  

Each episode independently samples a puddle center uniformly
along the track, $\mu_{f,\mathrm{dry}}\sim U(0.92,1.12)$, and
$\mu_{r,\mathrm{dry}}\sim U(0.97,1.19)$. The Gaussian puddle blends dry friction with $0.6$ as
$\mu=(1-w_{\mathrm{p}})\mu_{\mathrm{dry}}+w_{\mathrm{p}}\cdot0.6$,
with $w_{\mathrm{p}}=\exp[-\tfrac12(d_s^2/\sigma_s^2+d_e^2/\sigma_e^2)]$.
Here $d_s,d_e$ are longitudinal and lateral offsets from the
puddle center, with standard deviations $\sigma_s\sim U(5,10)$\,m and
$\sigma_e=4$\,m. Neither the actor nor MPC observes these randomized
parameters. 

\textbf{2) Real-world deployment:}\;
We evaluate the method on a 2019 Lexus LC~500 on an oval course.
Its onboard computer uses an NVIDIA RTX~4070~Ti Super GPU and an Intel
Xeon E-2278GE CPU. State estimates come from a GPS system.
We compare MPC nominal (manually-tuned fixed weights) against MPC SHAC
with puddles at corner entry or exit, using two trials per controller per
scenario.

As shown in Figure~\ref{fig:realworld}, \emph{MPC SHAC completes all four trials}, whereas \emph{the nominal MPC spins out in all four trials}
due to model mismatch from the water puddles, demonstrating robustness improvements from RL training with domain randomization. Across the four trials, max sideslip range is
$3.5^\circ$--$4.2^\circ$ for MPC SHAC versus $35^\circ$--$169^\circ$ for
MPC nominal; max absolute yaw rates are $0.77$--$0.85$ versus
$1.59$--$1.74$\,rad/s, respectively.

Results show that MPC SHAC brakes earlier on the corner approach, produces smoother
torques and steering inputs, and is able to maintain low sideslip throughout. On the initial puddle-free $38$\,m approach segment, the learned MPC is as fast as the nominal MPC. 
Over the $100\,\textrm{m}$ puddle-corner segment, MPC SHAC completes all passes in $6\,\textrm{s}$ with sideslip always below $3.2^\circ$, while nominal MPC’s completed passes have a mean peak absolute sideslip of $29.3^\circ$ and take $8\,\textrm{s}$ on average, demonstrating that RL training of MPC policies enables both faster and more robust racing performance.

Comparing the learned cost map with the  nominal MPC weights shows that the learned reference yaw-rate and sideslip $(r,\beta)$ tracking weights are about $100\times $ larger on average than the nominal ones. As a result, the learned MPC operates in a desired stable, low $(r,\beta)$ regime, so the policy avoids sliding and spinning out over puddles.  Simultaneously, the learned reference speed profile is about $2\,\textrm{m/s}$ faster on average, showing that RL training optimizes  for both faster racing performance and robustness through  domain randomization.

\FloatBarrier

\vspace{-1pt}
\section{Conclusion and Future Work}
\label{sec:conclusion}
We presented \solverName{}, a smoothed SQP--ADMM solver and
FORL pipeline for efficient learning of MPC policies on GPUs.
Simulation and hardware experiments demonstrate higher training throughput and improved racing performance under
model mismatch. 
In future work we are interested in expanding our evaluation  to learn cost and dynamics parameters simultaneously.
This raises an open question: how do we balance MPC planning performance with the physical fidelity of the learned model?
Future work also envisions extending the approach to incorporate contact.

\vspace{-1pt}
\section*{Acknowledgments}
We used LLM tools, including Codex and Claude Code, to assist with software implementation, drafting, proofreading, and review. All final text, figures, code, and references were edited, revised, and reviewed by humans.

\bibliographystyle{IEEEtran}
\bibliography{references}

\appendices

\section{Quadratic Penalty Relaxation}
\label{app:inequality-relaxation}

The pure log-barrier formulation \eqref{eq:snlp} can be augmented
with a quadratically penalized inequality relaxation~\cite{admmslack}:
\begin{equation}
\label{eq:app-relaxed-snlp}
\begin{aligned}
\min_{x,\xi,s}\quad & \ell(x;\theta)
+ \tfrac{\gamma_\xi}{2}\|\xi\|^2
- \kappa\textstyle\sum_{i=1}^{p}\log s_i \\
\text{s.t.}\quad & f(x;\theta)=0,\\
&g(x;\theta)-\xi+s=h(\theta),
\qquad s>0 .
\end{aligned}
\end{equation}
Here $\xi$ relaxes the inequalities as the parameters vary during learning,
preventing inequality infeasibility from stalling training, and
$\gamma_\xi>0$ is its quadratic penalty. The pure-barrier formulation
\eqref{eq:snlp} is recovered as $\gamma_\xi\to\infty$.

Stationarity with respect to $s$ and $\xi$ gives
$y_g=\kappa/s$ and $\gamma_\xi\xi=y_g$, respectively. Eliminating
$\xi=\gamma_\xi^{-1}y_g$ yields
\begin{equation}
\label{eq:app-relaxed-snlp-kkt}
F^{\mathrm{NLP}}_{\kappa,\xi}(w;\theta) :=
\begin{bmatrix}
\nabla_x\ell + C^\top y_f + G^\top y_g\\[1pt]
f\\[1pt]
s \circ y_g - \kappa\ones
\end{bmatrix} = 0,
\end{equation}
\begin{equation}
\label{eq:app-relaxed-snlp-sdef}
s := h(\theta)-g(x;\theta)
+\gamma_\xi^{-1}y_g \;(>0),
\qquad y_g>0 .
\end{equation}
The block $D$ in \eqref{eq:adjoint} then becomes
$D=\diag(s+\gamma_\xi^{-1}y_g)$. The pure-barrier solution has local
$O(\kappa)$ primal error as $\kappa\to0^+$, assuming twice continuously
differentiable data, LICQ, SOSC, and strict complementarity at the solution
of \eqref{eq:nlp}~\cite[Theorem~3]{frey2025diffnmpc}.
The additional relaxation satisfies $\|\xi\|=\|y_g\|/\gamma_\xi$
exactly and vanishes as $\gamma_\xi\to\infty$ for bounded multipliers.

\compactsubsection{Slack-relaxed Smooth QP Subproblem}

Applying the same relaxation to the local SQP model
\eqref{eq:barrier-sub} gives
\begin{equation}
\label{eq:app-relaxed-barrier-sub}
\begin{aligned}
\min_{x,\xi,s}\quad
&\half x^\top Px+q^\top x
+\tfrac{\gamma_\xi}{2}\|\xi\|^2
-\kappa\textstyle\sum_{i=1}^{p}\log s_i\\
\text{s.t.}\quad
&Cx=c,\qquad
s=\bar h-Gx+\xi>0 .
\end{aligned}
\end{equation}
Its KKT system, after eliminating $\xi$, is
\begin{equation}
\label{eq:app-relaxed-pkkt}
F^{\mathrm{QP}}_{\kappa,\xi}(w;\theta) :=
\begin{bmatrix}
Px+q+C^\top y_f+G^\top y_g\\[1pt]
Cx-c\\[1pt]
s\circ y_g-\kappa\ones
\end{bmatrix}=0,
\end{equation}
\begin{equation}
\label{eq:app-relaxed-sdef}
s:=\bar h-Gx+\gamma_\xi^{-1}y_g\;(>0),
\qquad y_g>0 .
\end{equation}
Taking $\gamma_\xi\to\infty$ reduces
\eqref{eq:app-relaxed-barrier-sub}--\eqref{eq:app-relaxed-sdef} to
the subproblem \eqref{eq:barrier-sub} and its KKT system \eqref{eq:pkkt}.

\compactsubsection{Slack-relaxed SAQP Updates}

For the slack-relaxed subproblem \eqref{eq:app-relaxed-barrier-sub}, the operator
split in \eqref{eq:split-smooth} becomes
\begin{equation}
\label{eq:app-relaxed-split}
\begin{aligned}
\min_{x,z,\xi,s}\quad
&\half x^\top Px+q^\top x+\mathcal I_{\{c\}}(z_f)
+B_{\kappa,\xi}(\xi,s)\\
\text{s.t.}\quad
&Ax-z=0,\\
&s=\bar h-z_g+\xi>0,
\end{aligned}
\end{equation}
where
$B_{\kappa,\xi}(\xi,s):=\tfrac{\gamma_\xi}{2}\|\xi\|^2
-\kappa\sum_{i=1}^{p}\log s_i$.
The inequality-block proximal operator is, componentwise,
\begin{equation}
\label{eq:app-relaxed-zprox}
(z_g,\xi)^{k+1}
=\argmin_{z,\xi}\; -\kappa\log s+\tfrac{\gamma_\xi}{2}\xi^2
+\tfrac{\rho_g}{2}(z-\bar z_g^{k+1})^2,
\end{equation}
with $s=\bar h-z+\xi>0$. Stationarity gives
$\kappa/s=\rho_g(\bar z_g^{k+1}-z)$ and
$\gamma_\xi\xi=\kappa/s$. The resulting closed-form update is
\begin{equation}
\label{eq:app-relaxed-retr-update}
\begin{alignedat}{2}
\bar z_g^{k+1}&:=\hat z_g^{k+1}+\rho_g^{-1}y_g^k,
&\quad r^{k+1}&:=\bar h-\bar z_g^{k+1},\\
\Gamma&:=\kappa(\rho_g^{-1}+\gamma_\xi^{-1}),
&s^{k+1}&:=b_\Gamma(r^{k+1}),\\
\xi^{k+1}&:=\frac{\kappa}{\gamma_\xi s^{k+1}},
&z_g^{k+1}&:=\bar h-s^{k+1}+\xi^{k+1}.
\end{alignedat}
\end{equation}
For $\kappa=0$, let $\Pi(v)=\min(v,\bar h)$ denote componentwise
projection onto the upper-bound set. The corresponding update is
\begin{equation}
\label{eq:app-relaxed-projection}
\begin{aligned}
z_g^{k+1}
&:=\frac{\rho_g}{\gamma_\xi+\rho_g}\bar z_g^{k+1}
+\frac{\gamma_\xi}{\gamma_\xi+\rho_g}
\Pi(\bar z_g^{k+1}),\\
\xi^{k+1}
&:=\frac{\rho_g}{\gamma_\xi+\rho_g}
\bigl(\bar z_g^{k+1}-\Pi(\bar z_g^{k+1})\bigr).
\end{aligned}
\end{equation}

\compactsubsection{Slack-relaxed KKT Consistency}

At a converged SQP solution $w^*$, define
$s^*:=h(\theta)-g(x^*;\theta)+\gamma_\xi^{-1}y_g^*$.
By \eqref{eq:sqp-data}, the exact-Hessian QP-KKT residual
\eqref{eq:app-relaxed-pkkt} coincides with its NLP-KKT residual:
\begingroup
\large
\begin{equation}
\label{eq:app-relaxed-consistency}
\begin{gathered}
F^{\mathrm{QP}}_{\kappa,\xi}(w^*;\theta)
=F^{\mathrm{NLP}}_{\kappa,\xi}(w^*;\theta)\\[2pt]
=
\begin{bmatrix}
\nabla_x\ell(x^*;\theta)+C^\top y_f^*+G^\top y_g^*\\[2pt]
f(x^*;\theta)\\[2pt]
s^*\circ y_g^*-\kappa\ones
\end{bmatrix}=0.
\end{gathered}
\end{equation}
\endgroup
The reduced adjoint system \eqref{eq:folded} retains the same form, with
\begin{equation}
\label{eq:app-relaxed-weight}
W^{-1}:=\diag\!\left(
\frac{y_g}{s+\gamma_\xi^{-1}y_g}
\right).
\end{equation}

\compactsubsection{Bounded Inequality Sensitivity and Limiting Cases}

With slack relaxation enabled, the inequality-sensitivity map is uniformly
bounded:
\begin{equation}
\label{eq:app-Wbound}
\kappa>0
\quad\Longrightarrow\quad
0\preceq W^{-1}\preceq\gamma_\xi I.
\end{equation}
Hence $G^\top W^{-1}G\preceq\gamma_\xi G^\top G$, preventing unbounded
inequality scaling in the reduced adjoint system.

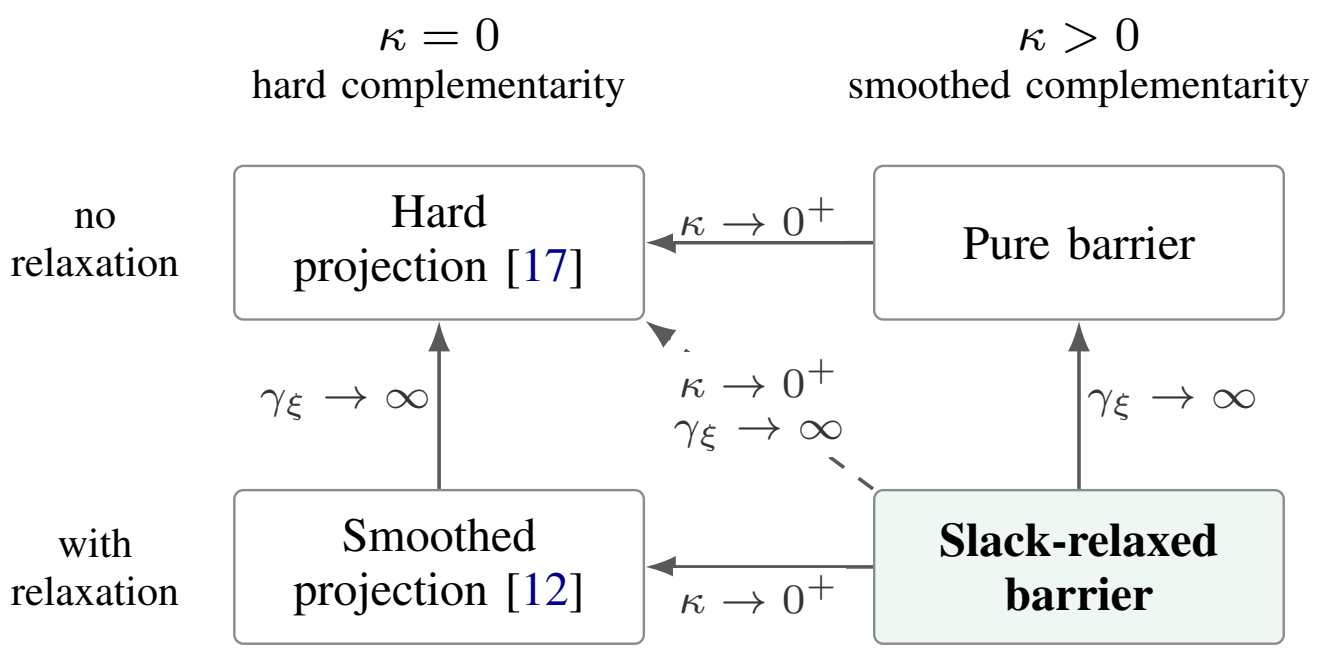
\begin{figure}[!t]
\centering
\resizebox{\columnwidth}{!}{\begin{tikzpicture}[
  formulation/.style={draw=black!45,rounded corners=1.5pt,align=center,
    minimum width=2.4cm,minimum height=0.9cm,inner sep=3pt,font=\footnotesize},
  limit/.style={-{Latex[length=1.8mm,width=1.3mm]},semithick,black!65},
  limitlabel/.style={font=\scriptsize,fill=white,inner sep=1pt,text=black!80}
]
\node[align=center,font=\footnotesize] at (-1.45,2.0)
  {$\kappa=0$\\[-1pt]{\scriptsize hard complementarity}};
\node[align=center,font=\footnotesize] at (2.3,2.0)
  {$\kappa>0$\\[-1pt]{\scriptsize smoothed complementarity}};
\node[anchor=east,align=center,font=\scriptsize] at (-2.85,0.95)
  {no\\relaxation};
\node[anchor=east,align=center,font=\scriptsize] at (-2.85,-0.95)
  {with\\relaxation};

\node[formulation] (hard) at (-1.45,0.95)
  {Hard\\projection~\cite{osqp}};
\node[formulation] (pure) at (2.3,0.95)
  {Pure barrier};
\node[formulation] (smooth) at (-1.45,-0.95)
  {Smoothed\\projection~\cite{turbompc}};
\node[formulation,fill=ForestGreen!5] (slack) at (2.3,-0.95)
  {\textbf{Slack-relaxed}\\\textbf{barrier}};

\draw[limit] (pure.west) -- node[limitlabel,above] {$\kappa\to0^+$} (hard.east);
\draw[limit] (slack.west) -- node[limitlabel,below] {$\kappa\to0^+$} (smooth.east);
\draw[limit] (smooth.north) -- node[limitlabel,left] {$\gamma_\xi\to\infty$} (hard.south);
\draw[limit] (slack.north) -- node[limitlabel,right] {$\gamma_\xi\to\infty$} (pure.south);
\draw[limit,dashed] (slack.north west) --
  node[limitlabel,align=center] {$\kappa\to0^+$\\[-1pt]$\gamma_\xi\to\infty$}
  (hard.south east);
\end{tikzpicture}}
\caption{Relationships among the four inequality treatments.}
\label{fig:app-smoothing-family}
\end{figure}

The limit $\kappa\to0^+$ recovers $s\circ y_g=0$. For bounded $y_g$,
the relaxation term $\gamma_\xi^{-1}y_g$ vanishes as
$\gamma_\xi\to\infty$. Consequently, any finite limit of solutions to
\eqref{eq:app-relaxed-snlp-kkt}--\eqref{eq:app-relaxed-snlp-sdef}
as $\kappa\to0^+$ and $\gamma_\xi\to\infty$ satisfies
the original NLP-KKT conditions \eqref{eq:nlp-kkt}. Likewise,
$b_\Gamma(v)\to\max(v,0)$ makes the slack-relaxed log-barrier update
\eqref{eq:app-relaxed-retr-update} recover the projection update
\eqref{eq:app-relaxed-projection}.

\compactsubsection{Effect on Example~1}
We repeat Example~1 with inequality relaxation in the unsmoothed baseline,
using $\gamma_\xi=10^3$. This stiff penalty approximates the hard NLP
while retaining its active-set kinks.

\begin{figure}[H]
  \centering
  \includegraphics[width=0.485\columnwidth]
    {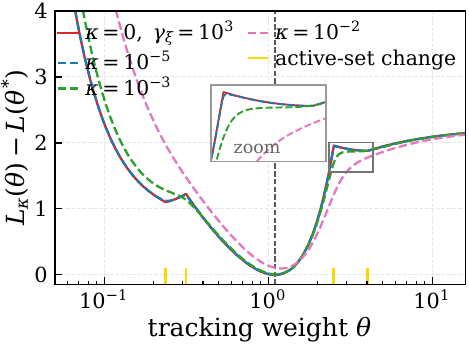}\hfill
  \includegraphics[width=0.485\columnwidth]
    {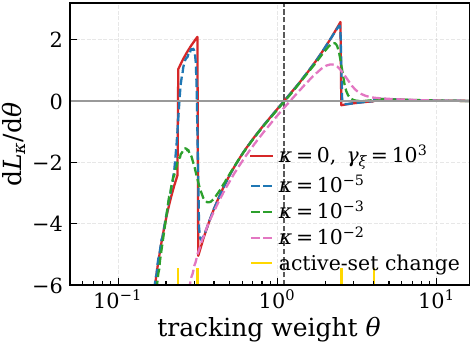}\\[-2pt]
  \makebox[0.485\columnwidth]{\scriptsize (a) Slack-relaxed outer loss}\hfill
  \makebox[0.485\columnwidth]{\scriptsize (b) Slack-relaxed gradient}\\[2pt]
  \includegraphics[width=0.485\columnwidth]
    {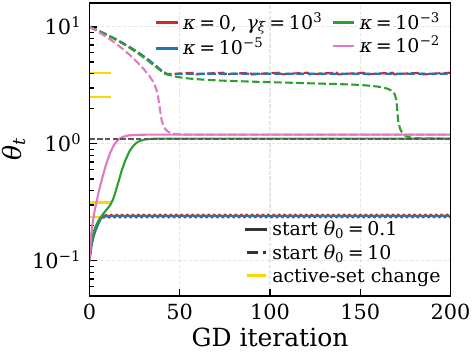}\hfill
  \includegraphics[width=0.485\columnwidth]
    {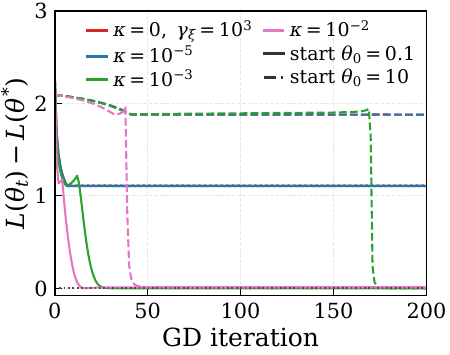}\\[-2pt]
  \makebox[0.485\columnwidth]{\scriptsize (c) GD iterates}\hfill
  \makebox[0.485\columnwidth]{\scriptsize (d) Learning progress}
  \caption{Double-integrator parameter learning from
  $\theta_0\in\{0.1,10\}$ with inequality relaxation enabled in the
  unsmoothed baseline ($\gamma_\xi=10^3$). The
  slack-relaxed NLP still stalls at active-set kinks, whereas sufficient
  barrier smoothing reaches $\theta^\star=1.10$ from both initializations.}
  \label{fig:app-bilevel-toy-slack}
\end{figure}

\begin{figure}[H]
  \centering
  \includegraphics[width=0.485\columnwidth]
    {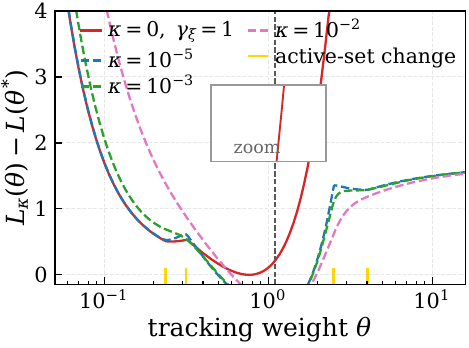}\hfill
  \includegraphics[width=0.485\columnwidth]
    {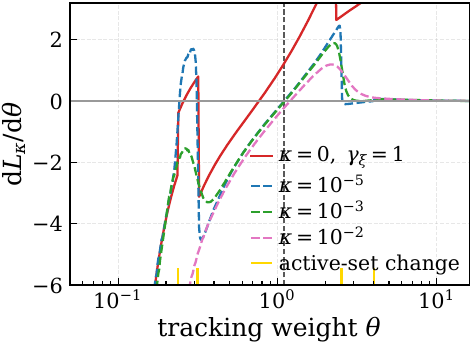}\\[-2pt]
  \makebox[0.485\columnwidth]{\scriptsize (a) Slack-relaxed outer loss}\hfill
  \makebox[0.485\columnwidth]{\scriptsize (b) Slack-relaxed gradient}\\[2pt]
  \includegraphics[width=0.485\columnwidth]
    {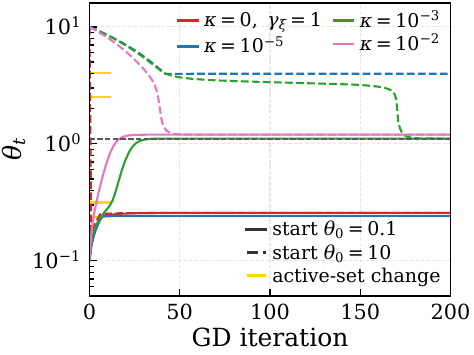}\hfill
  \includegraphics[width=0.485\columnwidth]
    {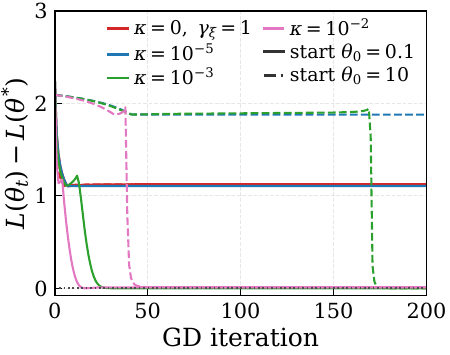}\\[-2pt]
  \makebox[0.485\columnwidth]{\scriptsize (c) GD iterates}\hfill
  \makebox[0.485\columnwidth]{\scriptsize (d) Learning progress}
  \caption{Double-integrator parameter learning from
  $\theta_0\in\{0.1,10\}$ with a soft inequality-relaxation penalty
  $\gamma_\xi=1$. Both initializations converge on the relaxed surrogate,
  but the learned parameter is biased when deployed on the original hard
  NLP. The barrier-smoothed curves are shown for comparison.}
  \label{fig:app-bilevel-toy-slack-soft}
\end{figure}

For comparison, Fig.~\ref{fig:app-bilevel-toy-slack-soft} uses the softer
penalty $\gamma_\xi=1$. The relaxed surrogate deviates from
the hard NLP: both initializations converge to a biased parameter whose
deployment on the original hard problem remains suboptimal.

\section{Two-Sided Inequality Constraints}
\label{app:two-sided}

We extend \texttt{SAQP} to two-sided constraints
$$h_l(\theta)\le g(x;\theta)\le h_u(\theta).$$ For brevity, we assume that
both bounds are finite as the one-sided formulation in
Sec.~\ref{sec:method} recovers the other cases. Linearizing as in
\eqref{eq:sqp-data} gives $$\bar h_l\le Gx\le\bar h_u,$$ with
$\bar h_{l}:=h_{l}(\theta)-g(x^j;\theta)+Gx^j$ and $\bar h_u$ defined
analogously. Each constraint carries the barrier
$$\phi(z):=-\kappa\log(z-\bar h_l)-\kappa\log(\bar h_u-z).$$

\subsection{Two-Sided Proximal Update}

The slack update \eqref{eq:zprox} becomes,
\begin{equation}
\label{eq:app-two-sided-zprox}
z_g^{k+1}=\argmin
\phi(z)+\tfrac{\rho_g}{2}(z-\bar z_g^{k+1})^2,
\end{equation}
with $\nu_l:=z-\bar h_l$ and $\nu_u:=\bar h_u-z$. Stationarity requires the
root of
\begin{equation}
\label{eq:app-two-sided-root}
r(z):=z-\bar z_g^{k+1}+\Gamma\Bigl(\frac{1}{\nu_u}-\frac{1}{\nu_l}\Bigr)=0,
\end{equation}
where $\Gamma=\kappa\rho_g^{-1}$ and its derivative  is given by
\begin{equation}
r'(z)=1+\frac{\Gamma}{\nu_l^2}+\frac{\Gamma}{\nu_u^2}>0 .
\end{equation}
Since $r\to-\infty$ as $z\to\bar h_l^+$ and $r\to\infty$ as
$z\to\bar h_u^-$, the root is unique and lies strictly inside the bounds.
By the implicit function theorem, the update is smooth in
$\bar z_g^{k+1}$ with derivative $1/r'\in(0,1)$. The dual update
\eqref{eq:admm-dual} gives
$y_g^{k+1}=\rho_g(\bar z_g^{k+1}-z_g^{k+1})=y_u-y_l$, with
$y_l:=\kappa/\nu_l$ and $y_u:=\kappa/\nu_u$. Hence
$\nu_l y_l=\nu_u y_u=\kappa$ after every ADMM iteration, generalizing
Remark~\ref{rmk:complementarity}. As $\kappa\to0^+$, the update tends to
$\min(\max(\bar z_g^{k+1},\bar h_l),\bar h_u)$, the box projection
of~\cite{osqp}.

\subsection{Safeguarded Newton Iteration}

Eq.~\eqref{eq:app-two-sided-root} is a cubic, which we solve with a fixed
number of Newton steps, keeping the update branch-free and of constant cost
on the GPU. At the midpoint $m:=(\bar h_l+\bar h_u)/2$, $\nu_l=\nu_u$, so
the barrier terms cancel and $r(m)=m-\bar z_g^{k+1}$. Since $r$ is strictly
increasing, the root lies on the same side of $m$ as $\bar z_g^{k+1}$.
Hence, the nearer bound dominates the barrier, and we warm-start from its
one-sided solution,
\begin{equation}
\label{eq:app-two-sided-warm}
z^0=\begin{cases}
\min\bigl(\bar h_l+b_\Gamma(\bar z_g^{k+1}-\bar h_l),\,m\bigr),
& \bar z_g^{k+1}\le m,\\
\max\bigl(\bar h_u-b_\Gamma(\bar h_u-\bar z_g^{k+1}),\,m\bigr),
& \text{otherwise}.
\end{cases}
\end{equation}
Since $r$ is singular at the bounds, a full Newton step
$\Delta z=-r(z)/r'(z)$ may leave $(\bar h_l,\bar h_u)$. We therefore update
$z\leftarrow z+\alpha\Delta z$ with
\begin{equation}
\label{eq:app-two-sided-ftb}
\alpha=\begin{cases}
\min\bigl(1,\,\zeta\,\nu_l/|\Delta z|\bigr), & \Delta z<0,\\
\min\bigl(1,\,\zeta\,\nu_u/\Delta z\bigr), & \text{otherwise},
\end{cases}
\end{equation}
where $\zeta\in(0,1)$ is a fraction-to-boundary constant, so each step
covers at most a fraction $\zeta$ of the distance to the bound it
approaches.

\begin{figure}[!t]
\centering
\includegraphics[width=\columnwidth]{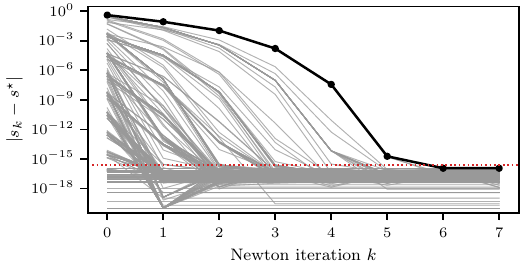}
\caption{Absolute error of the safeguarded Newton iterates. The black
line is the worst case over all test inputs and values of $\Gamma$.}
\label{fig:app-newton}
\end{figure}

We use seven Newton steps with $\zeta=0.99$. To test this choice, we
fix $[\bar h_l,\bar h_u]=[-1,1]$ and run the Newton iteration for a
distribution of log-barrier functions sampled in $\Gamma\in[10^{-20},10^{10}]$,
with inputs $\bar z_g^{k+1}$ inside, on, and up to $10^{8}$ outside
the bounds. Fig.~\ref{fig:app-newton} shows that seven steps
suffice for machine precision across all of these inputs.
The worst-case error is on the order of $10^{-15}$ after five steps and
reaches machine precision of the closest representable root after six.

\subsection{Inequality Relaxation and Sensitivities}

We next derive the slack update and implicit gradients for two-sided
constraints with inequality relaxation. Applying the quadratic penalty
from Appendix~\ref{app:inequality-relaxation} changes the slack update
\eqref{eq:app-two-sided-zprox} to
$$(z_g,\xi)^{k+1}=\argmin_{z,\xi}\;
\phi(z-\xi)+\tfrac{\gamma_\xi}{2}\xi^2
+\tfrac{\rho_g}{2}(z-\bar z_g^{k+1})^2.$$
Substituting $w:=z-\xi$ and minimizing over $\xi$ in closed form gives
$\xi=\tfrac{\rho_g}{\rho_g+\gamma_\xi}(\bar z_g^{k+1}-w)$.
After eliminating $\xi$, the update reduces to
$$
w^{k+1}=\argmin_w\;
\phi(w)+\frac{\tilde\rho_g}{2}(w-\bar z_g^{k+1})^2,
$$
where $\tilde\rho_g:=\rho_g\gamma_\xi/(\rho_g+\gamma_\xi)$
is the effective quadratic penalty coefficient.
The barrier $\phi$ and the center $\bar z_g^{k+1}$ of the quadratic
term are unchanged from \eqref{eq:app-two-sided-zprox}; eliminating
$\xi$ only replaces $\rho_g$ by $\tilde\rho_g$.
We therefore compute $w^{k+1}$ using the same safeguarded Newton
iteration, with
$\Gamma=\kappa/\tilde\rho_g
=\kappa(\rho_g^{-1}+\gamma_\xi^{-1})$,
and recover $z_g^{k+1}=w^{k+1}+\xi^{k+1}$.
With only an upper bound, the root has the
closed form $w^{k+1}=\bar h_u-b_\Gamma(\bar h_u-\bar z_g^{k+1})$, and the
update reduces to the one-sided update \eqref{eq:app-relaxed-retr-update}
with slack $s^{k+1}=\bar h_u-w^{k+1}$.

The backward pass combines the barrier curvatures of the lower
and upper bounds, $H_l:=\kappa/\nu_l^2$ and $H_u:=\kappa/\nu_u^2$, into
$H:=H_l+H_u$. With relaxation, we evaluate these curvatures using
$\nu_l=w-\bar h_l$ and $\nu_u=\bar h_u-w$. The reduced adjoint system
\eqref{eq:folded} retains its form, with inequality weights
\begin{equation}
\label{eq:app-two-sided-weight}
W^{-1}=\diag(H)\quad\text{and}\quad
W^{-1}=\diag\!\Bigl(\frac{\gamma_\xi H}{\gamma_\xi+H}\Bigr)
\end{equation}
without and with relaxation, respectively. For a single upper bound,
$H=\kappa/s^2$ and $y_g=\kappa/s$. Each diagonal entry of the relaxed
weight matrix then reduces to $y_g/(s+\gamma_\xi^{-1}y_g)$, recovering
the one-sided weight \eqref{eq:app-relaxed-weight} in
Appendix~\ref{app:inequality-relaxation}.

For parameter gradients, we include derivatives of both bounds. The
inequality term in \eqref{eq:grads} becomes
\begin{equation}
\label{eq:app-two-sided-grads}
\Bigl[\partial_\theta g-\tfrac{H_l}{H}\,\partial_\theta h_l
-\tfrac{H_u}{H}\,\partial_\theta h_u\Bigr]^\top\bar\lambda_g.
\end{equation}
If only an upper bound $h_u=h$ is present, then $H_l=0$ and $H_u=H$.
Equation~\eqref{eq:app-two-sided-grads} therefore reduces to
$[\partial_\theta(g-h)]^\top\bar\lambda_g$, the upper-bound term
derived in \eqref{eq:grads} of Sec.~\ref{sec:implicit-gpu}.

\FloatBarrier
\section{Racing Training Details}
\label{app:training-details}

\subsection{Learning Comparison}
\label{app:comparison-training}

For the comparison in Sec.~\ref{sec:exp-training}, \solverName{} uses
32 spatial intervals at $3$\,m spacing and one SQP iteration per
$50$\,Hz control step. We warm-start from the previous solution and use
merit-based step acceptance. The barrier parameter is $\kappa=10^{-4}$,
the inequality-relaxation penalty is $\gamma_\xi=10^5$, and the SAQP
iteration limit is 500. All methods run on a single NVIDIA RTX~5090.

The actor and critic each use two hidden layers of 256 units. All learned
policies use Adam~\cite{kingma2015adam}, $B=1024$ parallel environments,
discount $\gamma=0.99$, $\lambda=0.95$, and critic learning rate $10^{-3}$.
PPO uses ten optimization epochs and a clipping threshold of $0.2$.
Table~\ref{tab:hyper} lists the settings that differ across methods.
Every ten updates, we evaluate each policy without exploration for
1000 control steps from the start line.

\begin{table}[H]
\caption{Hyperparameters for the learning comparison
(Sec.~\ref{sec:exp-training}). $^\dagger$Learning rate annealed to zero.}
\label{tab:hyper}
\centering
\footnotesize
\setlength{\tabcolsep}{3pt}
\begin{tabular}{@{}lrrrr@{}}
\toprule
 & \multicolumn{2}{c}{MPC} & \multicolumn{2}{c@{}}{Direct}\\
\cmidrule(lr){2-3}\cmidrule(l){4-5}
 & SHAC & PPO & SHAC & PPO\\
\midrule
Rollout steps & $32$ & $128$ & $32$ & $128$\\
Updates & $100$ & $100$ & $400$ & $100$\\
Actor learning rate & $5\!\cdot\!10^{-4}$ & $10^{-5}$ & $2.5\!\cdot\!10^{-6}$\,$^\dagger$ & $10^{-4}$\,$^\dagger$\\
Minibatches/epoch & -- & $256$ & -- & $32$\\
\bottomrule
\end{tabular}
\end{table}

\FloatBarrier
\subsection{Deployment Training}
\label{app:deployment-training}

For deployment in Sec.~\ref{sec:exp-robustness}, we train an MPC policy
from scratch with SHAC on a single NVIDIA RTX~5090, using double precision
throughout. Table~\ref{tab:deployment-training} summarizes the training
settings. The simulator uses two fixed Tsit5 steps of $0.01$\,s per
$0.02$\,s control interval. Episodes continue across SHAC rollouts until
termination or the time limit.

\begin{table}[t]
\caption{Hyperparameters for deployment training
(Sec.~\ref{sec:exp-robustness}).}
\label{tab:deployment-training}
\centering
\footnotesize
\setlength{\tabcolsep}{3pt}
\begin{tabular}{@{}p{0.43\columnwidth}p{0.53\columnwidth}@{}}
\toprule
Setting & Value\\
\midrule
Parallel environments & $B=512$\\
Control / simulation step & $0.02$\,s / $0.01$\,s\\
MPC horizon & $32\times3$\,m $=96$\,m\\
SQP / SAQP iteration limits & 1 / 500\\
SHAC rollout horizon & $T=128$ steps ($2.56$\,s)\\
Maximum episode & 1500 steps ($30$\,s)\\
Training updates & 250\\
Discount / TD($\lambda$) & $\gamma=0.995$, $\lambda=0.95$\\
Actor learning rate & $5\times10^{-4}$ to $10^{-4}$, linearly decayed\\
Critic learning rate & $10^{-3}$\\
Optimizer & Adam, $\beta=(0.9,0.99)$\\
Gradient norm cap & $1.0$\\
Critic fitting per update & 16 passes, 4 minibatches each\\
Target critic EMA decay & $0.995$\\
Exploration standard deviations & $(\sigma_\delta,\sigma_{\tau_r},\sigma_{\tau_f})$ $=(0.0025,5,10)$\\
\bottomrule
\end{tabular}
\end{table}

\textbf{Reference.}
The MPC and RL reward use the same reference lap of length $L=258.5$\,m,
with a $10\%$ faster speed profile.

\textbf{Randomization.}
At the start of each episode, we randomize the initial state and the
simulated friction conditions. We sample the initial position uniformly
along the lap and add independent uniform perturbations to the
corresponding reference state. The perturbations in
$(r,v,\beta,\omega_r,e,\Delta\phi)$ have symmetric bounds
$\pm(0.05,0.25,0.01,0.25,0.05,0.02)$, respectively.
We sample the front and rear dry-friction coefficients independently and
uniformly within $\pm10\%$ of their nominal values, $(1.02,1.08)$, and
sample the center of a Gaussian puddle uniformly along the lap.
The puddle reduces friction toward $0.6$, with longitudinal spread
$\sigma_s\sim U(5,10)$\,m and lateral spread $\sigma_e=4$\,m.
The friction coefficients and puddle parameters remain fixed within each
episode. The MPC retains nominal vehicle parameters, and neither the
actor nor the MPC observes the friction or puddle parameters.

\textbf{Policy parameterization.}
The actor and critic each have two hidden layers of 256 ReLU units. The
actor has 38 outputs, and the critic predicts a scalar value. Their
45-dimensional observation contains seven scaled vehicle states, a
sine--cosine encoding of lap position, 33 curvature samples at $3$\,m spacing, and three
scaled previous controls. Running cost weights are shared across the MPC horizon.

The actor outputs quadratic cost weights and bounded offsets that
parameterize the linear state cost coefficients $q_\theta$ and
$q_\theta^H$ introduced in Sec.~\ref{sec:exp-robustness}. For both running
and terminal costs, we compute these coefficients as
$q=-2Q\,\delta x_{\mathrm{ref}}$. Here $\delta x_{\mathrm{ref}}$ shifts
the preferred tracking target in the MPC objective through the linear
cost term. The offsets in speed, rear-wheel speed, lateral deviation,
and relative heading satisfy $|\delta x_{\mathrm{ref},i}|\le b_i$, where
$(b_v,b_{\omega_r},b_e,b_{\Delta\phi})=(5,5,1,0.2)$. All other offsets
are zero.

The quadratic weight on time is zero. In the racing cost function
(Sec.~\ref{sec:exp-training}), we use the fixed control-rate matrix
$W=\diag(10^4,10^{-5},10^{-9})$ and terminal-time weight $w_t=100$.

\textbf{Reward and termination.}
The reward encourages progress along the track and penalizes tracking
errors and violations of the speed, track, and heading limits.
For a transition with finite numerical values,
\begin{equation}
\label{eq:deployment-reward}
r_t=\Delta s_t-\tfrac{20}{6}\sum_{j=1}^{6}
\left(\frac{\varepsilon_{j,t+1}}{a_j}\right)^2
-500\bigl[\mathcal B(x_{t+1})+\mathbf1_{\mathrm{failure}}\bigr],
\end{equation}
where $\Delta s_t$ is signed progress along the track in meters.
The vector $\varepsilon_{t+1}$ contains tracking errors in
$(r,v,\beta,\omega_r,e,\Delta\phi)$, and
$a=(5,25,2,25,5,2)$ gives their normalization scales.
These errors are measured against the fixed
reference, so the learned MPC offsets do not change the reward's
tracking target.

Let $e_{\min}$ and $e_{\max}$ denote the lateral track boundaries.
The soft penalty $\mathcal B$ is
\begin{equation}
\begin{aligned}
\mathcal B(x)={}&\max\!\left(0,\frac{5-v}{30}\right)^2\\
&+\max\!\left(0,\frac{e_{\min}+0.15-e}{5}\right)^2\\
&+\max\!\left(0,\frac{e-e_{\max}+0.15}{5}\right)^2\\
&+\max\!\left(0,|\Delta\phi|-\frac{\pi}{3}\right)^2.
\end{aligned}
\end{equation}
The lateral penalties begin $0.15$\,m inside each track boundary,
so they apply before the vehicle leaves the track.

Episodes end after 1500 control steps ($30$\,s), or earlier if:
\begin{itemize}[leftmargin=3mm,itemsep=0pt,parsep=0pt,topsep=2pt]
\item Speed falls below $5$\,m/s.
\item The vehicle leaves the track: $e<e_{\min}$ or $e>e_{\max}$.
\item The heading error satisfies $|\Delta\phi|>\pi/3$.
\item A transition produces NaN or infinite values.
\end{itemize}
The indicator $\mathbf1_{\mathrm{failure}}$ equals one for the first
three conditions and zero otherwise. Nonfinite transitions receive
$r_t=-500$.
\textbf{Optimization.}
We add Gaussian action noise with the standard deviations in
Table~\ref{tab:deployment-training}.
Rewards are normalized by a running standard deviation of discounted
returns, held fixed within each rollout, and clipped to $\pm1000$.
Critic targets use normalized rewards, and the actor objective is rescaled
to the original reward units.

\end{document}